\documentclass[journal]{IEEEtran}
\usepackage{amsmath,amsfonts}
\usepackage{algorithmicx}
\usepackage{algorithm}
\usepackage{array}
\usepackage[caption=false,font=normalsize,labelfont=sf,textfont=sf]{subfig}
\usepackage{textcomp}
\usepackage{stfloats}
\usepackage{url}
\usepackage{verbatim}
\usepackage{graphicx}
\usepackage{cite}
\usepackage{hyperref}
\usepackage{color}
\usepackage{xcolor}
\usepackage{graphicx}
\usepackage{booktabs}
\usepackage{amsmath, amsthm, amsfonts, amssymb}
\usepackage{mathrsfs}
\usepackage{textcomp}
\usepackage{epstopdf}
\usepackage{multirow}
\usepackage{wrapfig}
\usepackage{subfig}
\usepackage{booktabs} 

\usepackage{algpseudocode}
\usepackage{ragged2e}
\usepackage[normalem]{ulem}
\usepackage{cleveref}
\usepackage{bm}

\definecolor{green}{rgb}{0, 0.5, 0}
\definecolor{orange}{rgb}{0.6, 0.3, 0.1}
\definecolor{red}{rgb}{1.0, 0.0, 0.0}
\definecolor{teal}{rgb}{0.0, 0.4, 0.4}
\definecolor{purple}{rgb}{0.65,0,0.65}
\definecolor{saffron}{rgb}{0.95,0.75,0.2}
\definecolor{turquoise}{rgb}{0.0,0.5,0.5}
\definecolor{brown}{rgb}{0.5, 0.16, 0.16}
\definecolor{brickred}{rgb}{.6, .2 .1}
\definecolor{coral}{rgb}{1,0.45,0.33}
\definecolor{newcolor}{rgb}{.8,.349,.1}
\definecolor{mygreen}{RGB}{15, 153, 5}
\definecolor{myorange}{RGB}{255, 153, 5}

\usepackage{tikz,color,hyperref}

\definecolor{lime}{HTML}{A6CE39}
\DeclareRobustCommand{\orcidicon}{
\begin{tikzpicture}
\draw[lime, fill=lime] (0,0)
circle[radius=0.16]
node[white]{{\fontfamily{qag}\selectfont \tiny \.{I}D}}; 
\end{tikzpicture}
\hspace{-2mm}
}
\foreach \x in {A, ..., Z}{%
\expandafter\xdef\csname orcid\x\endcsname{\noexpand\href{https://orcid.org/\csname orcidauthor\x\endcsname}{\noexpand\orcidicon}}
}

\begin{document}
\title{Motion-Adapter: A Diffusion Model Adapter for Text-to-Motion Generation of Compound Actions}

\author{Yue Jiang,
        Mingyu Yang,
        Liuyuxin Yang,
        Yang Xu,
        Bingxin Yun,
        and Yuhe Zhang*
        
\thanks{Yue Jiang, Mingyu Yang, Liuyuxin Yang, Yang Xu, Bingxin Yun and Yuhe Zhang are with the School of Computer Science, Northwest University, Xi'an, China. Yuhe Zhang is the corresponding author. E-mail: zhangyuhe0601@nwu.edu.cn.}

\thanks{Manuscript submitted Apr 16, 2026}}

\maketitle

\begin{abstract}

Recent advances in generative motion synthesis have enabled the production of realistic human motions from diverse input modalities. However, synthesizing compound actions from texts, which integrate multiple concurrent actions into coherent full-body sequences, remains a major challenge. We identify two key limitations in current text-to-motion diffusion models: (i) catastrophic neglect, where earlier actions are overwritten by later ones due to improper handling of temporal information, and (ii) attention collapse, which arises from excessive feature fusion in cross-attention mechanisms. As a result, existing approaches often depend on overly detailed textual descriptions (\textit{e.g.}, raising right hand), explicit body-part specifications (\textit{e.g.}, editing the upper body), or the use of large language models (LLMs) for body-part interpretation. These strategies lead to deficient semantic representations of physical structures and kinematic mechanisms, limiting the ability to incorporate natural behaviors such as greeting while walking. To address these issues, we propose the Motion-Adapter, a plug-and-play module that guides text-to-motion diffusion models in generating compound actions by computing decoupled cross-attention maps, which serve as structural masks during the denoising process. Extensive experiments demonstrate that our method consistently produces more faithful and coherent compound motions across diverse textual prompts, surpassing state-of-the-art approaches.
\end{abstract}


\maketitle

\section{Introduction}

\begin{figure*}[h]
  \includegraphics[width=\textwidth]{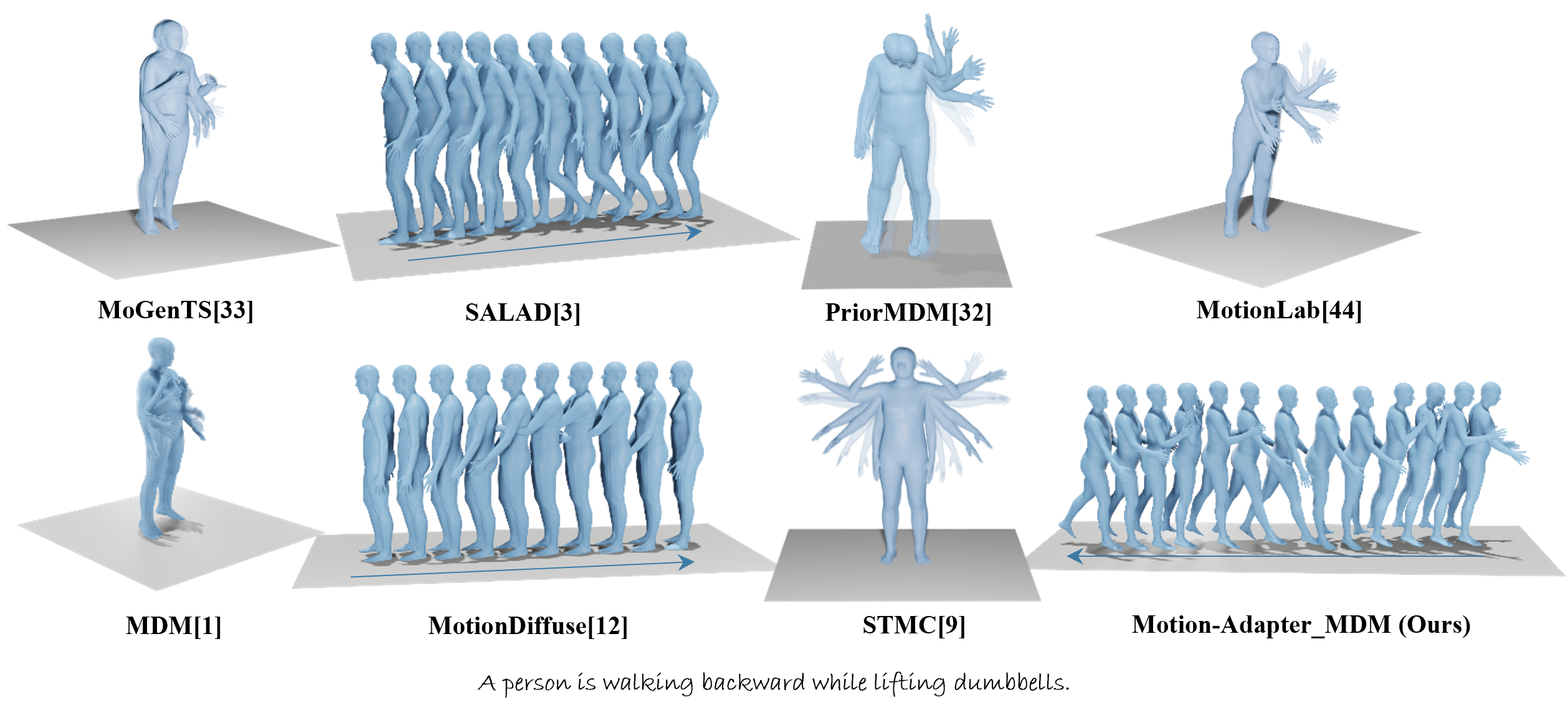}
  \caption{Motion sequences generated by our Motion-Adapter given a textual prompt and a pre-trained motion diffusion model (\textit{e.g.}, MDM~\cite{HMDM}). The Motion-Adapter\_MDM more accurately captures compound actions compared with existing models. Later frames are rendered with increased transparency for clarity, and arrows indicate the direction of displacement.} 
  \label{fig: teaser}
\end{figure*}

Recent progress in generative motion synthesis has enabled realistic human motion and stylized human motion generation from diverse inputs such as action labels~\cite{Action2Motion}, text~\cite{SALAD, SmooGPT}, and 3D skeletal data~\cite{Mo2024Motion}.  Building on this, we focus on generating motions with compound actions—integrating multiple sub-actions into coherent, full-body sequences. This setting better reflects real-world behaviors, where actions often co-occur (\textit{e.g.}, \textit{walking while lifting dumbbells}, see Figure~\ref{fig: teaser}) rather than unfold in simple succession.

Existing approaches for compound action generation often rely on detailed textual descriptions that explicitly specify body-part movements (\textit{e.g.}, 'raising the right hand')~\cite{TEMOS, DiffusionPriors} or are interpreted via large language models~\cite{SINC, STMC}. However, even LLM-derived descriptions frequently lack semantic precision: articulating actions such as 'snapping' may require verbose phrasing, while expressions like 'repeatedly reaching hands forward' fail to convey the specificity of 'punching'. Motion-editing frameworks that adopt a two-stage process~\cite{SALAD} offer an alternative, and part-based editing provides more targeted control. Nonetheless, these approaches introduce additional complexity by requiring separate encoders, sophisticated fusion mechanisms~\cite{MMM, SALAD}, or source motions as editing references~\cite{Motionfix}, which increases annotation demands and computational overhead, ultimately limiting scalability and efficiency.

\begin{figure}[ht]
    \begin{minipage}[b]{0.5\textwidth}
        \centering
        \includegraphics[width=\textwidth]{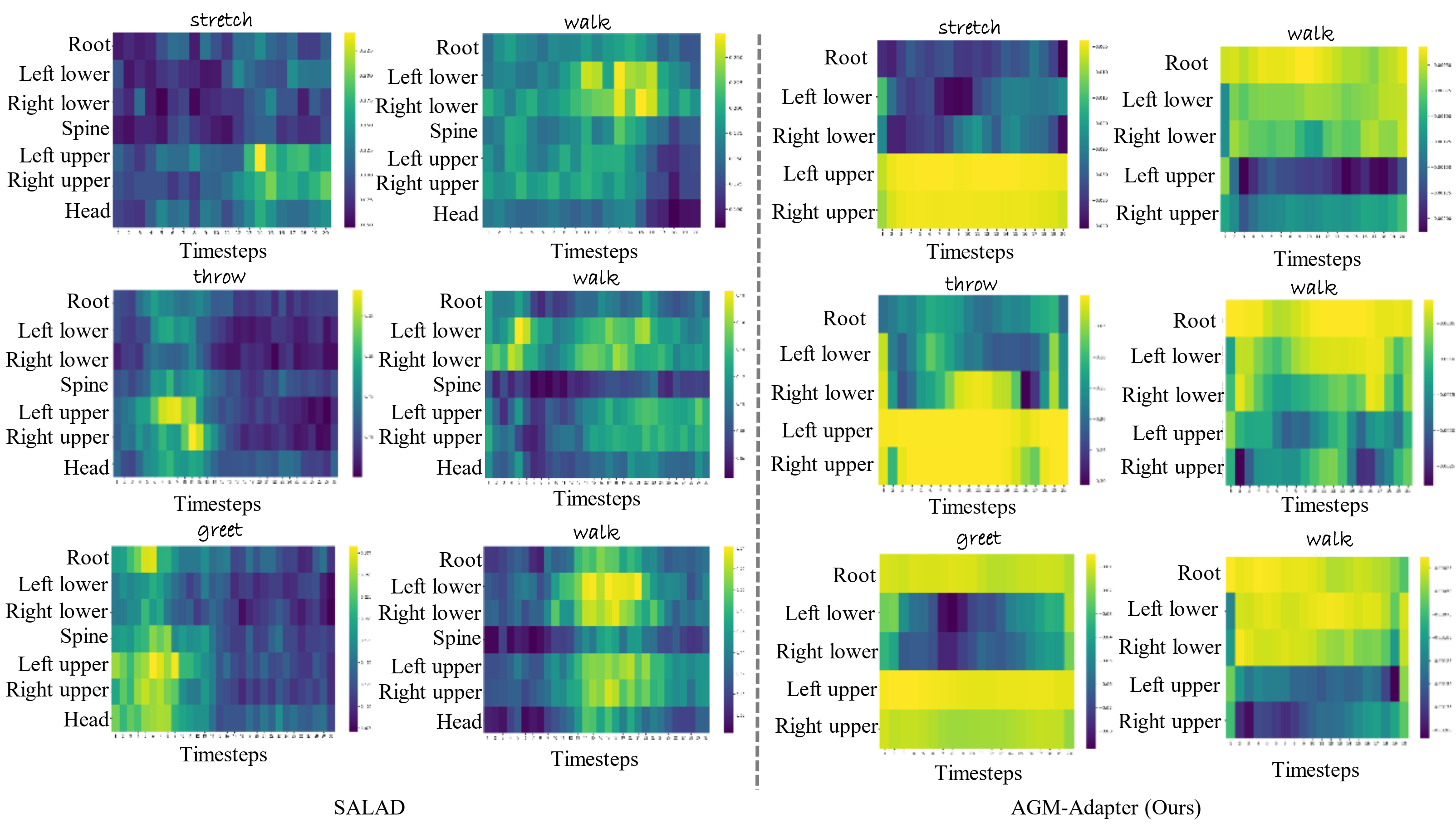}
        \caption{Comparison of attention maps from SALAD~\cite{SALAD} and our Motion-Adapter. SALAD's maps are averaged across all Transformer layers at step 40, while ours are taken from the third cross-attention layer at step 750.}
        \label{fig: attentionMap}
    \end{minipage}
\end{figure}

Generating compound actions from action verbs requires more than identifying body-part specifications; models must also infer the correspondence between actions and articulated segments while maintaining accurate and globally coherent motion. Current text-to-motion diffusion models face two key challenges: (i) catastrophic neglect, where earlier actions are overwritten during temporal fusion and decoding; and (ii) attention collapse, where post-feature fusion degrades spatial specificity. This collapse is common in motion generation pipelines, where the need to integrate body segments often leads to overly aggressive feature fusion, undermining the discriminative power of the original attention maps. For instance, as shown in Figure~\ref{fig: attentionMap}, SALAD~\cite{SALAD} localizes actions like throw and greet early on, and stretch and walk later, yet its attention remains diffusely spread across all joints. This makes interpreting or manipulating attention maps infeasible in models lacking cross-attention (\textit{e.g.}, MDM~\cite{HMDM}), over-fusing body parts (\textit{e.g.}, MotionDiffuse~\cite{MotionDiffuse}), or applying strong post-attention fusion (\textit{e.g.}, SALAD's FiLM layer). 

Therefore, we propose Motion-Adapter, a plug-and-play module that guides text-to-motion diffusion models in generating compound actions. It integrates cross-attention with convolutional layers to compute decoupled cross-attention maps, thereby inferring the body parts corresponding to each verb token. These cross-attention maps, serving as structural masks, are applied to noisy motions at each denoising step, guiding the generation toward semantically precise and structurally coherent actions. Motion-Adapter is trained solely on single-action motions and requires no modification or additional tuning of the diffusion backbone, enabling efficient and consistent motion synthesis.

We compared our framework with seven existing methods, including text-to-motion models, text-driven motion editing models, and spatial motion composition models. Extensive experiments show that our approach not only outperforms existing methods in generating compound actions and long-term motions, but also substantially improves the fidelity and diversity of the generated motions.

\section{Related Works}

\textbf{\textit{Action-to-Motion Generation}.} Early approaches primarily relied on discrete action labels as input to generate corresponding motion sequences that represent specific actions. Representative methods in this line of work include Action2Motion \cite{Action2Motion}, ACTOR \cite{ACTOR}, GL-Transformer \cite{GL-Transformer}, CVAE-based methods \cite{CVAE-based}, UM-CVAE \cite{zhong2022umcvae}, and PoseGPT \cite{TM2T}, which explore various generative frameworks such as conditional variational autoencoders and transformer-based architectures to capture the complex mapping between textual descriptions and motion data. Some of these approaches are capable of generating long-term motion sequences that comprise multiple actions, \textit{i.e.}, temporally sequential pose combinations—they struggle to produce compound action motions, often failing to produce motions that are both temporally coherent and semantically consistent.

\textbf{\textit{Text-to-Motion Generation}.} Text-conditioned motion generation aims to synthesize human motion sequences based on natural language descriptions. Action labels inherently lack fine-grained information like motion trajectories and steps, limiting the diversity and expressiveness of the generated motions. To address this limitation, several works have focused on generating motions conditioned on full textual descriptions, including Language2Pose \cite{Language2Pose}, TM2T \cite{gu2022tm2t}, a two-stage method \cite{TwoStage}, MMM \cite{MMM}, Transfer learning-based method \cite{Transfer-learning-based}, MoMask \cite{MoMask}, EMDM \cite{EMDM}, FlowMDM \cite{FlowMDM}, MotionMamba \cite{MotionMamba}, MotionDiffuse \cite{MotionDiffuse}, and MotionFlux\cite{MotionFlux}. These methods are capable of synthesizing motions conditioned on full textual descriptions, accurately generating action sequences that align with the input sentence. They can also produce long-term motion sequences composed of multiple connected actions and also support editing specific body parts using action labels.

Another line of research focuses on motion generation conditioned on full textual descriptions, which not only synthesizes motion sequences but also allows the specification of particular limbs (\textit{e.g.}, the left arm or right leg) to perform designated actions. Representative works in this category include TEMOS \cite{TEMOS}, MotionCLIP \cite{MotionCLIP}, MDM \cite{HMDM}, DNO \cite{DiffusionPriors}, and SALAD \cite{SALAD}.

\textbf{\textit{Text-driven Motion Editing}.} Text-driven motion editing typically includes style editing, body-part editing, and domain-specific pose editing. Among these, body-part editing is most relevant to our work. Many text-to-motion models such as MDM~\cite{HMDM}, MotionDiffuse~\cite{MotionDiffuse}, and SALAD~\cite{SALAD}—support this task by specifying the target body parts for text-based modification. Similarly, FLAME~\cite{FLAME2023} relies on explicit part selection. Recent methods like CoMo~\cite{CoMo} and FineMoGen~\cite{FineMoGen} leverage LLMs to generate edit instructions and show promising results in part-level editing. MotionFix~\cite{Motionfix}, while capable of editing body parts without explicit part specification, requires a triplet dataset of source motion, target motion, and corresponding edit text.

\textbf{\textit{Text-driven Motion Composition}.} The motion composition task can be broadly categorized into temporal composition, spatial composition, or a combination of both. Several work focuses on temporal composition, with the goal of producing long-term motions, such as TEACH\cite{TEACH}, PriorMDM\cite{priorMDM}, MoGenTS\cite{MoGenTS}. However, spatial composition, which aims to generate motions involving multiple actions occurring simultaneously, is more relevant to our work. Representative work includes SINC\cite{SINC}, MCD\cite{MCD}, EnergyMoGen\cite{EnergyMoGen} and STMC\cite{STMC}. The key distinction lies in the fact that the former three approaches require paired motion training data, and methods such as SINC \cite{SINC}, MCD \cite{MCD}, and STMC \cite{STMC} rely on large language models (LLMs) to decompose textual prompts and identify the referenced body parts. In contrast, our method requires neither paired training data nor explicit textual interpretation.


\begin{figure}[H]
\centering
    \includegraphics[width=0.49\textwidth]{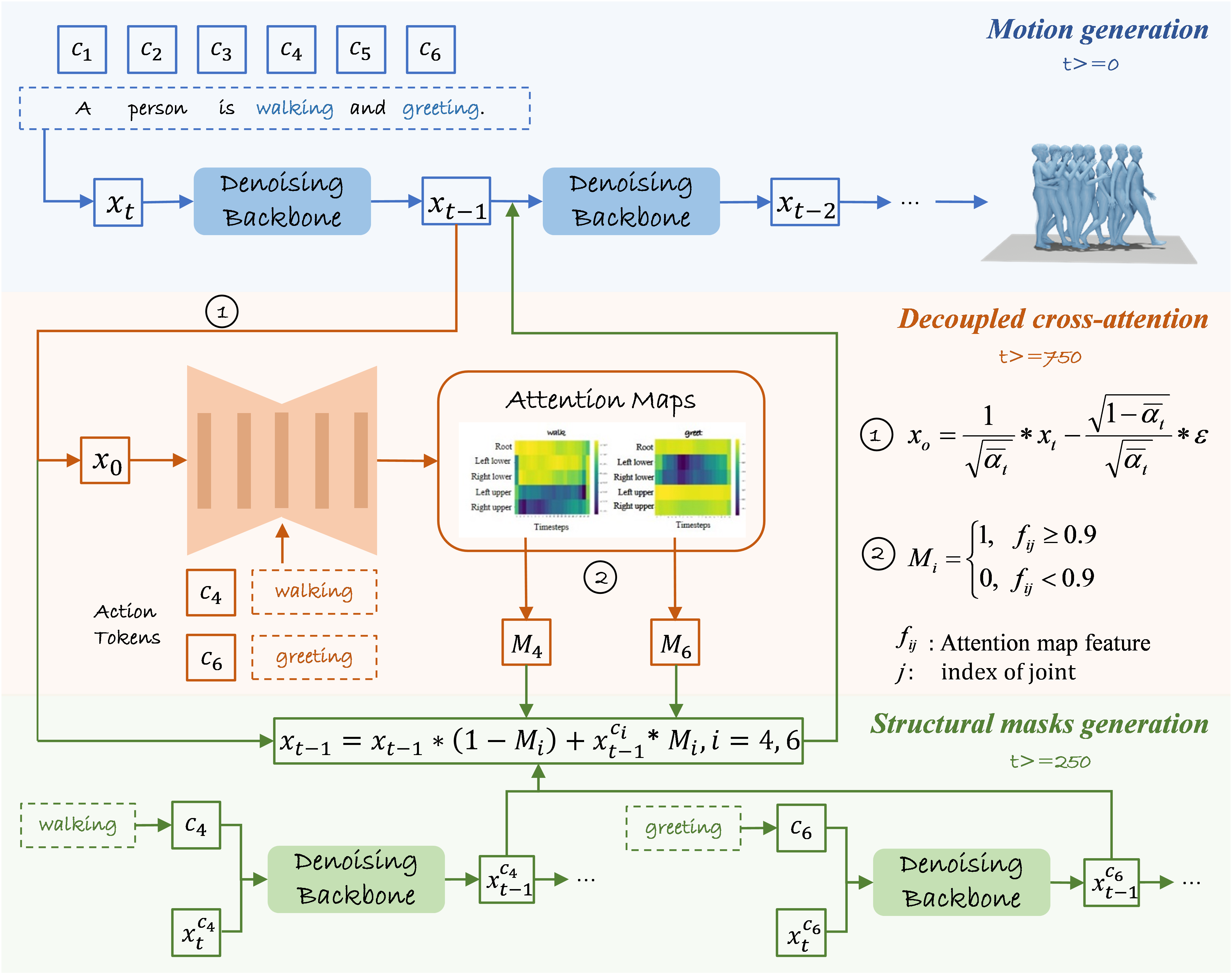}
    \caption{Overview of the Motion-Adapter integrated into the diffusion model at step $t$.}
    \label{fig: network}
\end{figure}

\section{Motion-Adapter}

In this paper, the Motion-Adapter computes decoupled cross-attention maps as structural masks to facilitate the generation of compound actions. As discussed in previous sections, existing methods struggle with compound action generation, primarily because effective cross-attention maps are difficult to extract as post-feature fusion often degrades spatial specificity within the attention mechanism. To address this limitation, we propose a decoupled cross-attention strategy, enabling the reliable extraction of cross-attention maps as structural masks. The overall architecture is illustrated in Figure~\ref{fig: network}. The Motion-Adapter comprises two main components: a network trained to extract decoupled cross-attention maps, and a structural mask generation module that embeds these masks into the pretrained text-to-motion diffusion model.





\subsection{Decoupled cross-attention} 



\begin{figure}[H]
\centering
    \includegraphics[width=0.5\textwidth]{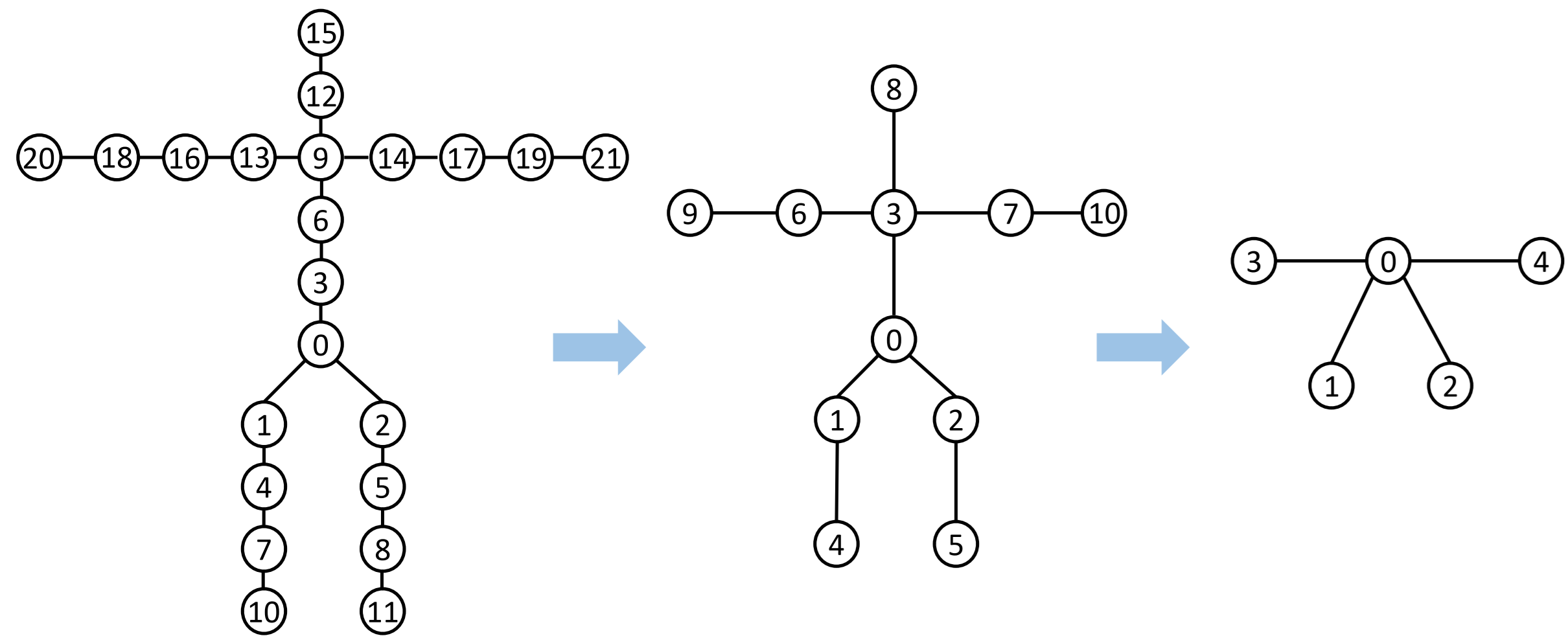}
    \caption{Illustration of the skeletal pooling on the HumanML3D dataset\cite{HumanML}.}
    \label{fig: body_parts}
\end{figure}

\textbf{\textit{Motion representation.}} Given a motion sequence $M$, we first decompose it into joint-wise features, denoted as $m_j \in \mathbb{R}^{N \times D_j}$, where $N$ is the number of frames and $D_j$ is the feature dimension for joint $j$, which varies according to the joint type. For example, in this work, we adopt the HumanML3D dataset \cite{HumanML}, where the skeleton consists of 22 joints (as shown in Figure \ref{fig: body_parts}), with $N = 196$ and $D_j = 3$, including with 3D spatial position.


\begin{figure}[t]
\centering
    \includegraphics[width=0.5\textwidth]{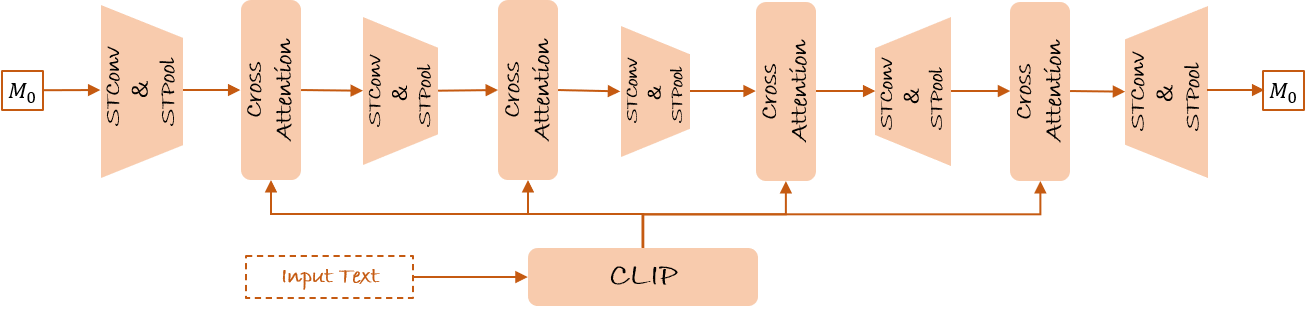}
    \caption{The architecture of the decoupled cross-attention.}
    \label{fig: networkSupp}
\end{figure}

\textbf{\textit{Network architecture.}} We propose a self-supervised framework composed of five STEncoder modules, each containing a STEncoder layer with an integrated cross-attention mechanism. The architecture is illustrated in Figure~\ref{fig: networkSupp}.

An STEncoder module~\cite{yan2018stgcn} consists of a STConv layer followed by a joint-pooling operation. The STConv layer integrates a 1D convolution for Skeletal Convolution (SkelConv) and a 1D convolution for Temporal Convolution (TempConv) to jointly capture spatial and temporal motion dynamics, while the pooling operation supports both downsampling and upsampling. In our design, joint features are treated as channels and expanded to varying feature dimensions across the five STConv modules. Along the temporal dimension, kernel sizes are applied with padding defined as $(\text{kernel size}-1)/2$ and a fixed stride of 2, thereby halving or doubling the temporal resolution at each stage. For the spatial dimension, convolution is performed by multiplying the joint–joint adjacency matrix with the motion feature matrix, which preserves spatial dimensionality. Down-sampling or up-sampling of joints is achieved exclusively through the joint-pooling operation. Specifically, in the first STConv module, max-pooling is applied along the skeletal dimension, reducing the number of joints from 22 to 11, as illustrated in Figure \ref{fig: body_parts}. The second STConv module further reduces the joint count to 5. Symmetrically, the fourth STEncoder module restores spatial resolution through alternating convolutional kernel size and upsampling, and the final convolutional layer maps the features back to the input dimensionality, producing the complete output motion sequence.

To incorporate semantic guidance, a cross-attention mechanism \cite{transfomer} is applied after each STEncoder module to align textual and motion representations. Text sequences are encoded with a pretrained CLIP model \cite{CLIP}, following the same protocol as MDM \cite{HMDM}. In this formulation, motion features act as queries, while textual features serve as keys and values, enabling information flow from text to motion. The cross-attention employs eight attention heads to capture fine-grained correspondences between textual tokens and skeletal joints. 

\textbf{\textit{Implementation details.}} The joint features are projected to dimensions of 64, 128, 256, 128, and 64, with convolutional kernel sizes of 9, 5, 3, 5, and 9 across the five STEncoder modules, respectively. The number of joints after pooling is 22, 11, 5, 11, and 22 for the corresponding modules.

We train Motion-Adapter in a self-supervised manner using single-action motions $M$ and their paired text as inputs, with $M$ serving as the reconstruction target. The loss is defined as the mean squared error (MSE) \cite{rumelhart1986learning} between the ground-truth motion $M$ and the predicted motion $\hat{M}$, indexed over the joints $j$, as shown in Equation (\ref{reconstruction_loss}).

\begin{equation} \label{reconstruction_loss}
\mathcal{L}_{\text{recon}} = \frac{1}{N} \sum_{j=1}^{N} \left\| \mathbf{m}_{j} - \hat{{m}_{j}} \right\|_2^2 ({m}_{j}\in M, \hat {m}_{j} \in \hat{M})
\end{equation}

Motion-Adapter was implemented on an NVIDIA 2080 Ti GPU and trained for 2,000 epochs on the HumanML3D dataset~\cite{HumanML} using the AdamW optimizer~\cite{Adam} with a batch size of 32 and a learning rate of 0.0001, without any additional labeling or paired motion data.

\begin{algorithm}
\caption{A Single Denoising Step using Motion-Adapter}
\label{Alg: Inference}
\begin{algorithmic}[1]
\Statex \hspace*{-\algorithmicindent} \textbf{Input:} A text prompt $c$, a set of action token indices $I$, the predicted result from the previous step $x_t$, a timestep $t$ and a pre-trained Diffusion Model.
\Statex \hspace*{-\algorithmicindent} \textbf{Output: }$x_{t-1}$

        \State  ${x}_{t-1}$, $\epsilon$  = DiffusionBackbone($c$, $x_t$)
        \State $x_0 = \frac{1}{\sqrt{\bar{\alpha}_t}} \cdot x_t - \frac{\sqrt{1 - \bar{\alpha}_t}}{\sqrt{\bar{\alpha}_t}} \cdot \epsilon$
        
        \State $F_{t-1}^{c}$ =  Motion-Adapter($c$, ${x_0}$)
        
        \For {$i \in I$} 
            \State $\text{$Mask_{t-1}^{c_i}$} \gets F_{t-1}^{c_i} $
            \State ${{{x}_{t-1}^{c_i}}}$ = DiffusionBackbone($c_i$, $x_t^{c_i}$) 
            
                
            \State ${x}_{t-1} = {{x}_{t-1}} * (1-Mask_{t-1}^{c_i}) + {{{x}_{t-1}^{c_i}}} * Mask_{t-1}^{c_i}$

        \EndFor
\State \Return $x_{t-1}$

\end{algorithmic}
\end{algorithm}

\subsection{Structural masks generation} 

We select the attention maps produced by the third cross-attention layer in the decoupled cross-attention, as they effectively represent body parts using a minimal number of joints. These maps are then transformed into structural masks, which are applied to $x_{t-1}$ tor guide the denoising process.  

Specifically, we extract the attention maps $F_t^{c_i}$ corresponding to each action token $c_i$ in the input sentence and normalize them to the range [0, 1], treating values above 0.9 as true activations. To enforce structural consistency, we introduce the following constraints: the two upper-body joints are grouped together, and since locomotion is primarily driven by the lower body, the root joint is bound to two lower-body joints. A region is considered active (TRUE) only if at least two joints from the upper or lower body are activated; otherwise, all associated joints are set to FALSE. This strategy helps prevent uncoordinated or anatomically implausible motion patterns. After aggregating the attention maps, we obtain the masks $Mask^{c_i}$ for each action token.

Given the masks $Mask^{c_i}$, $x_{t-1}$ is updated by combining each $Mask^{c_i}$ with $x_{t-1}^{c_i}$ according to Equation (\ref{maskEqu}), after which it is propagated to the subsequent denoising steps. 

\begin{equation} \label{maskEqu}
{x}_{t-1} = {{x}_{t-1}} * (1-Mask_{t-1}^{c_i}) + {{{x}_{t-1}^{c_i}}} * Mask_{t-1}^{c_i}
\end{equation}

\begin{table*}[htbp]
\centering
\caption{Upper-body and lower-body motion captions of our benchmark.}
\label{tab:body_motion_captions}

\begin{tabular}{ll}
\toprule
\textbf{Upper-body Motions} & \textbf{Lower-body Motions} \\
\midrule
is stretching & jogs in place\\
drinks something & kicks slowly\\
is lifting something & in a squatting position\\
is waving & sits down\\
drinks from the bottle & is running in place\\
is greeting & walks in place\\
claps in front of their body & is squatting down\\
is lifting a dumb bell & is doing jumping jacks in place\\
is eating a hamburger & walks in a curved line\\
looks at watch & walks in stride forward\\
picks up something & walking forward normally\\
is throwing something & does two small bunny hops\\
is raising & makes two short hops\\
throws an upper cut  & takes a small hop forwards\\
dodges & is standing as a subject\\

raises hands and crosses them & took four steps forward\\
stretches their arms side to side & is moving to the right\\
swats at something then protects his face with both hands & runs forward as if his feet hurt\\
raises their hands over their head and then down again & runs forward before stopping\\
is lifting & stumbles, then straightens up and continues walking\\
lifts up their hand as if fearful  & walks in a u shape\\
lifts both slightly  &runs in place at a very fast pace\\

\bottomrule
\end{tabular}
\end{table*}

\subsection{Training and inference}

Given the masks $Mask^{c_i}$ corresponding to each input action token $c_i$, optimization can be directly applied during the denoising process to generate compound action motions, without any parameter tuning, additional training data, or external conditioning.

A single denoising timestep $t$ is outlined in Algorithm~\ref{Alg: Inference}. At each step, an intermediate motion estimate ${x_0}$, is obtained and fed into the Motion-Adapter alongside the input text. The Motion-Adapter generates attention maps, which are subsequently transformed into the masks $Masks^{c_i}$. Crucially, using ${x_0}$ as the input to the Motion-Adapter, rather than the individual ${x^{c_i}_0}$, allows the network to mitigate misleading effects from training sequences that are neither purely static nor purely dynamic. For instance, the action 'throw' predominantly involves upper-body motion but is often sampled concurrently with locomotion (\textit{e.g.}, throwing while walking) in the dataset. Feeding ${x^{c_i}_0}$ individually into the Motion-Adapter can thus produce suboptimal attention maps. By contrast, using ${x_0}$ enables the network to more appropriately associate lower-body movements with corresponding lower-body actions and upper-body movements with upper-body actions, yielding soft guidance that enhances the robustness of the proposed method. Next, we apply each mask $Mask^{c_i}$ to the individual motion ${x_{t-1}^{c_i}}$ using Equation~(\ref{maskEqu}), and assign the results to $x_{t-1}$ to obtain a refined motion representation. This ensures that only the action-aligned body parts are updated with their corresponding values. 


We observe that after step $t = 750$, the attention maps $F_t$ no longer align well with the motion, which disrupts the generation process.To address this, we stop generating masks beyond $t = 750$ and continue using the results obtained up to this point. Furthermore, to preserve coherent and coordinated motion, we cease applying masks after $t = 250$. While strong activations on upper- and lower-body joints can help the model learn discriminative features, they may compromise overall coordination. Disabling such emphasis beyond $t = 250$ encourages smoother and more continuous motion.


Our Motion-Adapter produces masks that enable the soft integration of multiple sub-actions into unified, body-part-aligned motions, ensuring both semantic consistency and spatial coherence. Moreover, these masks support controllable editing across diverse motion diffusion models by operating on a shared motion representation structure. Detailed results are presented in Section~\ref{sec: results}.



\section{Results}
\label{sec: results}

To evaluate the Motion-Adapter, we integrate it with two diffusion backbones, MDM~\cite{HMDM} and MotionDiffuse~\cite{MotionDiffuse}, resulting in two variants: Motion-Adapter\_MDM and Motion-Adapter\_MotionDiffuse. We compare these variants against seven state-of-the-art methods, including text-to-motion models, text-driven motion editing models, and spatial composition models, under textual prompts of varying complexity. For visual clarity, we first present the comparison results of Motion-Adapter\_MDM against baseline text-to-motion models, followed by evaluations on spatial composition methods and text-driven motion editing. Finally, we report the results of Motion-Adapter\_MotionDiffuse on the same set of prompts in Section \ref{sec: MotionDiffuseResults}.

\subsection{Benchmark} 

To systematically evaluate compound action generation, we construct a benchmark based on the HumanML3D test set~\cite{HumanML}, as existing datasets do not explicitly address this task. The benchmark includes 44 action prompts—22 upper-body and 22 lower-body—detailed in Table~\ref{tab:body_motion_captions}. Upper-body prompts comprise 17 atomic verbs and 5 fine-grained descriptions, while lower-body prompts feature 22 compound actions with trajectories or adverbial modifiers. Pairing each upper-body prompt with each lower-body prompt yields 484 unique textual prompts. All ground-truth motions are generated by purely concatenating upper- and lower-body motions using forward kinematics principles~\cite{kinematics}, and no explicit mechanism ensures their coordination.

\subsection{Qualitative evaluation on simple prompts}
\label{sec: simplePrompts}

\begin{figure}[htb]
        \centering
        \includegraphics[width=0.45\textwidth]{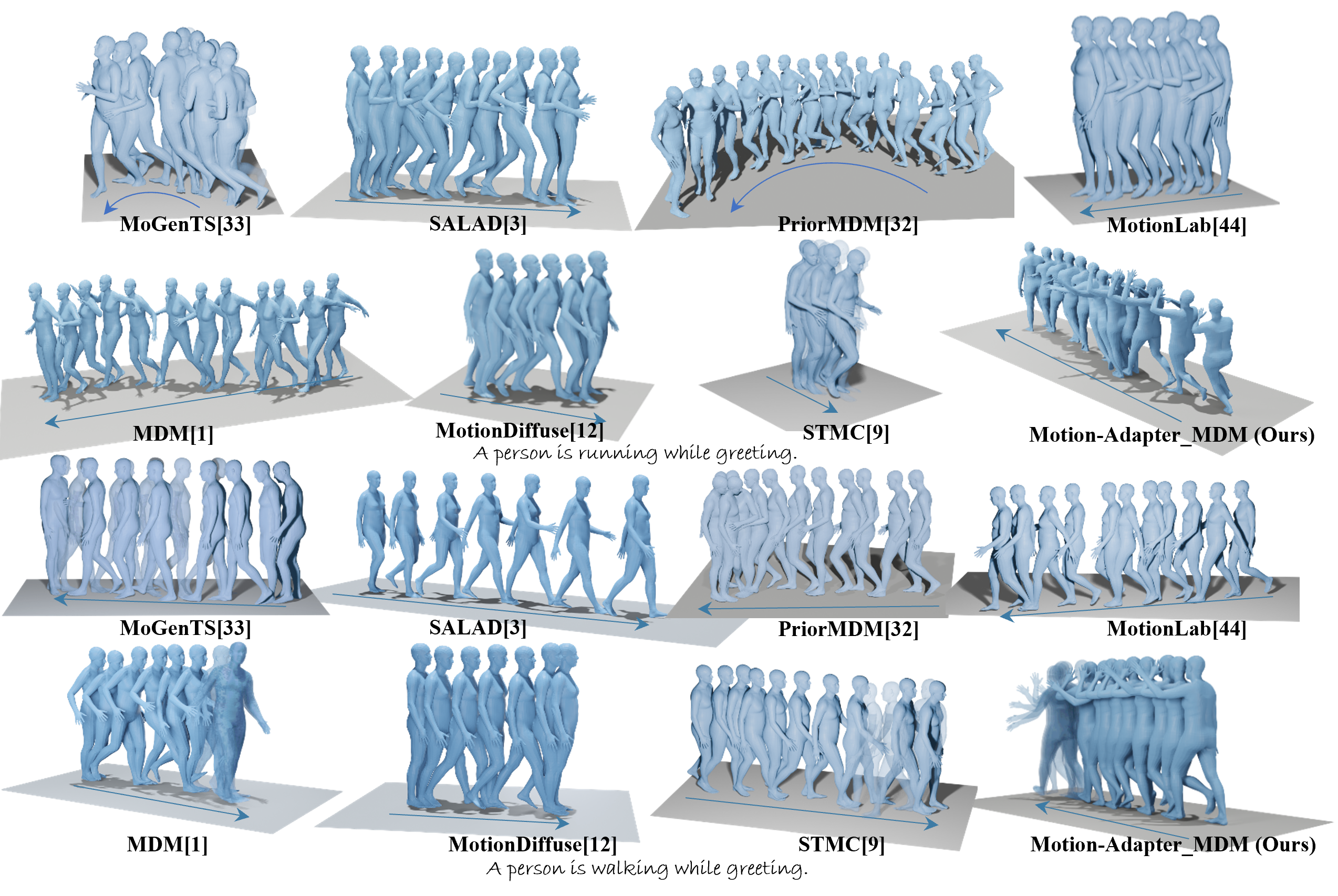}
        \caption{Qualitative comparison of compound actions combining 'greeting' with 'walking' or 'running'. Later frames are rendered with increased transparency for clarity, and arrows indicate motion direction.}
        \label{fig: greeting}
\end{figure}

\begin{figure}[htb]
        \centering
        \includegraphics[width=0.5\textwidth]{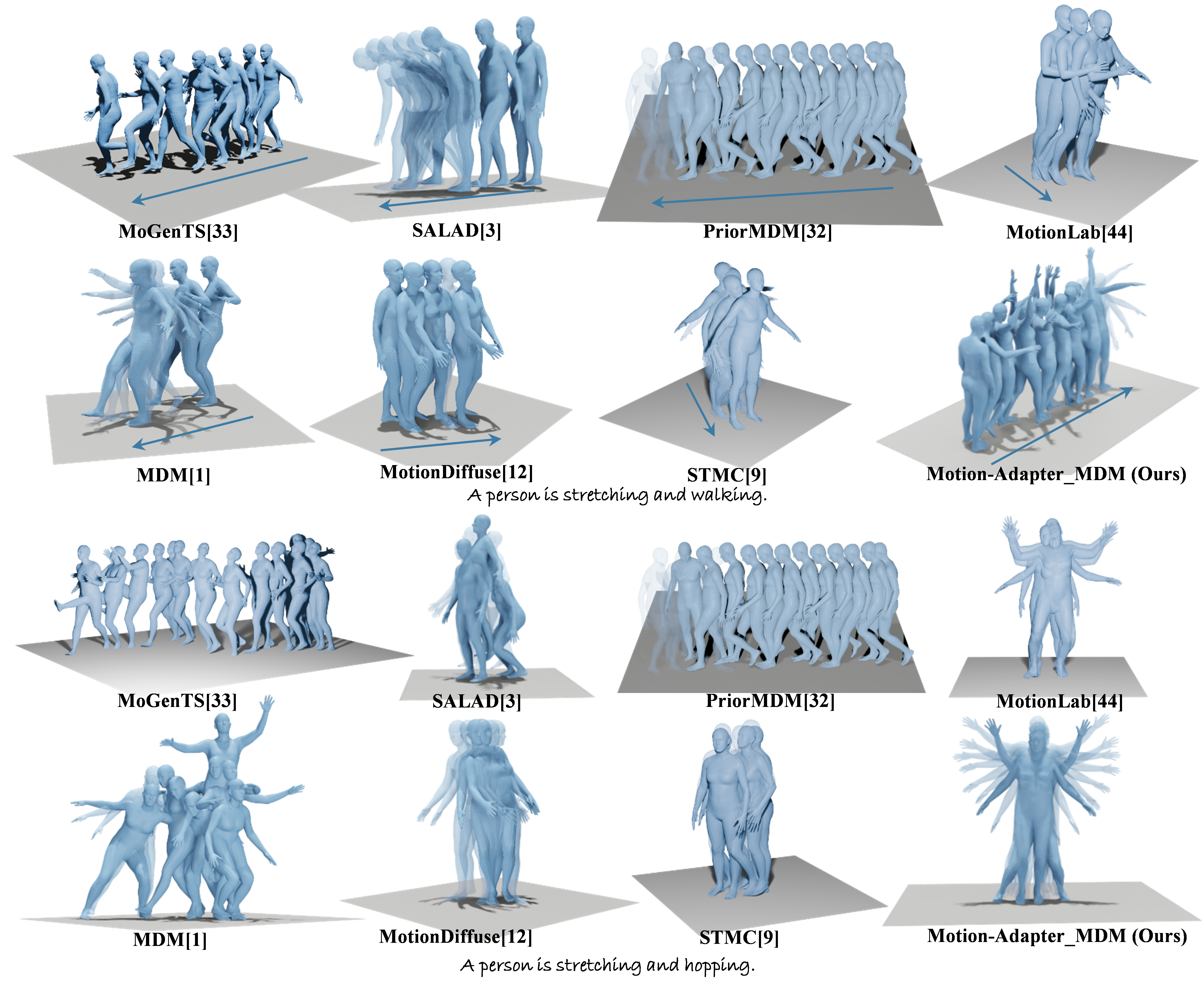}
        \caption{Qualitative comparison results of compound actions combining 'stretching' with either 'hopping' or 'walking'. Later frames are rendered with increased transparency for clarity, and arrows indicate motion direction.}
        \label{fig: stretching}
\end{figure}

\begin{figure}[htb]
        \centering
        \includegraphics[width=0.5\textwidth]{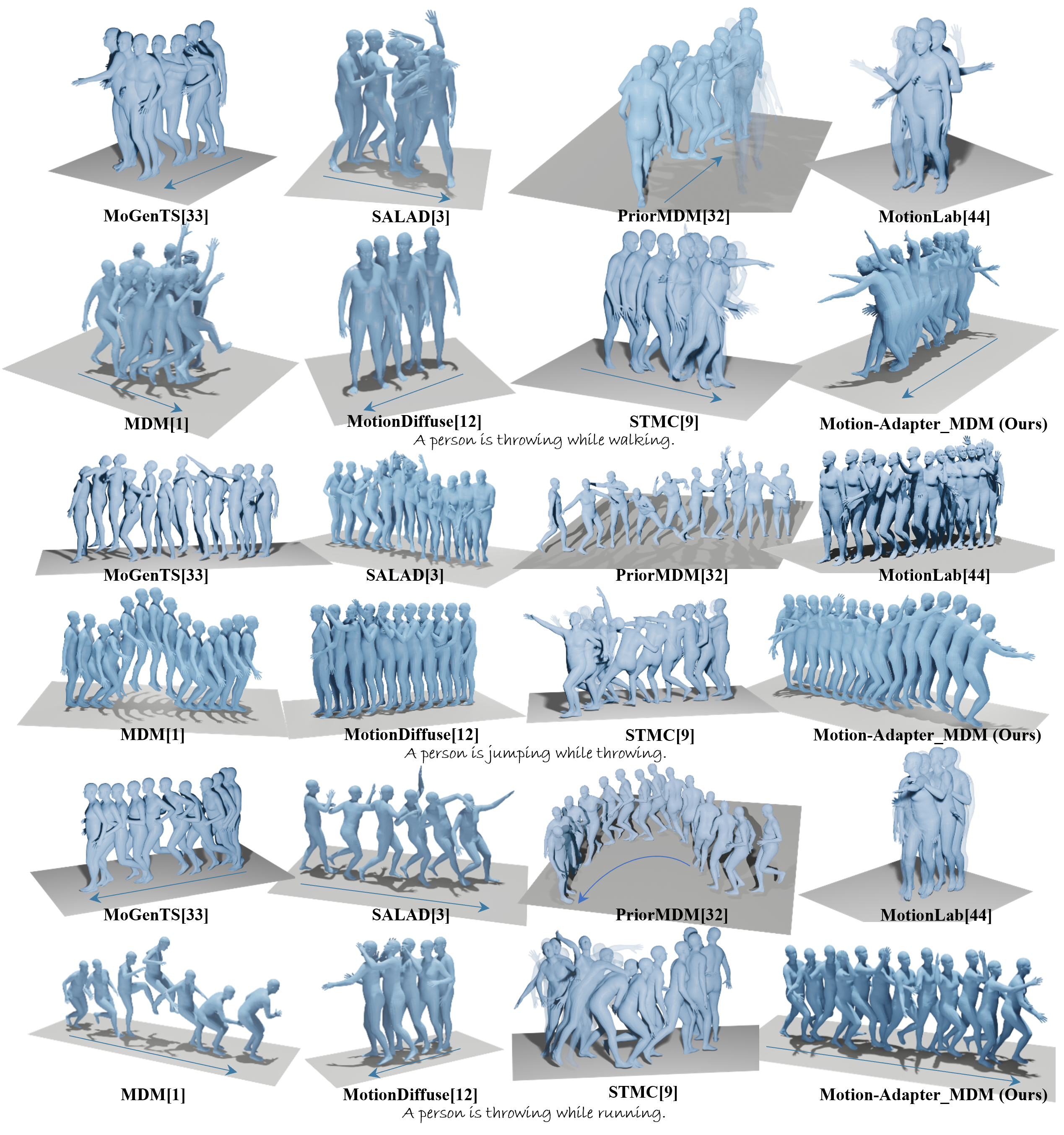}
        \caption{Qualitative comparison results of compound actions combining 'throwing' with either 'walking', 'running' or 'jumping'. Later frames are rendered with increased transparency for clarity, and arrows indicate motion direction.}
        \label{fig: throwing}
\end{figure}

\textbf{\textit{Comparison to text-to-motion models}.} In Figures \ref{fig: greeting}, \ref{fig: stretching} and \ref{fig: throwing}, we present the results for compound actions comprising two sub-actions\cite{loper2015smpl}. The first group combines 'greeting' with either 'walking' or 'running'; the second merges 'stretching' with either 'hopping' or 'walking', and the third pairs 'throwing' with 'walking', 'running', or 'jumping'.

As illustrated, our Motion-Adapter\_MDM successfully generates compound actions that are both natural and temporally coherent, whereas baseline models struggle to produce such motions. For instance, MoGenTS~\cite{MoGenTS} handles 'walking' and 'running' well but neglects simultaneous hand movements like 'greeting' or 'throwing'. SALAD~\cite{SALAD} relies heavily on the semantics of the first word in the action description, while MotionLab~\cite{Motionlab} tends to produce actions with minimal trajectories, performing poorly for movements such as 'walking' and 'running'. When attempting to combine actions, these methods can produce distorted body configurations, as seen in examples where PriorMDM~\cite{priorMDM} and MDM~\cite{HMDM} attempt to generate 'throwing and walking'.  Moreover, MotionDiffuse~\cite{MotionDiffuse} typically generates only a single action, such as 'walking' without 'stretching' or 'throwing' without 'jumping'.

\textbf{\textit{Comparison to spatial composition models}.} SINC~\cite{SINC}, MCD~\cite{MCD}, and EnergyMoGen~\cite{EnergyMoGen} are representative spatial composition models. However, as their code and pretrained models are unavailable, we compare our method only with STMC~\cite{STMC}, the most recent work in this area.

As shown in Figures \ref{fig: greeting}, \ref{fig: stretching} and \ref{fig: throwing}, STMC\cite{STMC} often omits one of the sub-actions (\textit{e.g.}, generating 'hopping' without 'stretching'), produces them sequentially (\textit{e.g.}, 'walking' followed by 'throwing'), or misidentifies the target actions, such as generating 'walking' followed by 'bending' instead of 'stretching', thereby diminishing the overall realism of the motion. This is because LLMs are statistical models trained on text corpora and primarily capture linguistic regularities rather than embodied semantics. Consequently, when describing body movements, they often fail to encode the semantic details of specific body parts or their motion dynamics, even though they can correctly identify the corresponding body parts. In contrast, our Motion-Adapter directly links semantic information to joint features through convolution and cross-attention mechanisms, effectively aligning motions with the corresponding body-part movements, while ensuring that the generated motions remain physically consistent.

\begin{figure}[htb]
        \centering
        \includegraphics[width=0.5\textwidth]{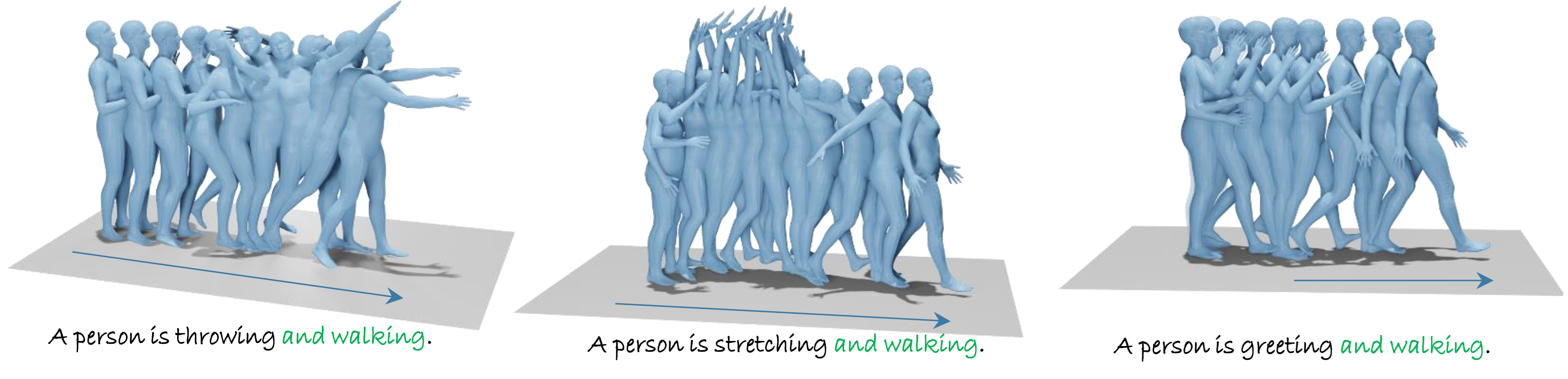}
        \caption{Motion editing results of SALAD\cite{SALAD}. The black text represents the source content, while the green text indicates the editing instructions.}
        \label{fig: motion_edit}
\end{figure}

\textbf{\textit{Comparison to text-driven motion editing model}.} In the context of motion editing, several methods such as MDM~\cite{HMDM}, SALAD~\cite{SALAD}, and MMM~\cite{MMM} support body-part-specific manipulation to generate compound actions. However, MDM\cite{HMDM} and MMM\cite{MMM} primarily focus on user-driven control over specific regions, typically the upper body, within a given source motion. It constitutes a simple motion concatenation rather than a synthesis that accounts for coordination and coherence. Among existing methods, the motion editing strategy termed 'prompt refinement' in SALAD~\cite{SALAD} is most relevant to our task. This two-stage approach first generates a single-action motion, then automatically identifies the editable regions to synthesize compound actions. For fair comparison, we evaluate our method against SALAD~\cite{SALAD} by constructing bidirectional action pairs: using 'greet', 'stretch', and 'throw' as base actions and adding 'walk' or 'run', and vice versa. The best results are shown in Figure~\ref{fig: motion_edit}, with quantitative comparisons in Table~\ref{tab: Quatitative}. While SALAD~\cite{SALAD} succeeds in combinations such as 'stretching and walking' and 'throwing and walking', it also produces sequential motions such as greeting followed by walking, rather than truly integrated compound actions. 

\textbf{\textit{Summary.}} As observed from Figures \ref{fig: greeting}, \ref{fig: stretching} and \ref{fig: throwing}, our method maintains coherence and temporal consistency even in challenging cases such as 'greeting and running' or 'throwing and running', where both actions involve overlapping body regions (\textit{e.g.}, arms). This demonstrates that generating compound actions requires more than identifying or bridging specific body parts and underscores the effectiveness and robustness of our Motion-Adapter in handling such complex motions.

\subsection{Qualitative evaluation on complex prompts}
\label{sec: complexePrompts}

\begin{figure*}[tb]
        \centering        \includegraphics[width=1\textwidth]{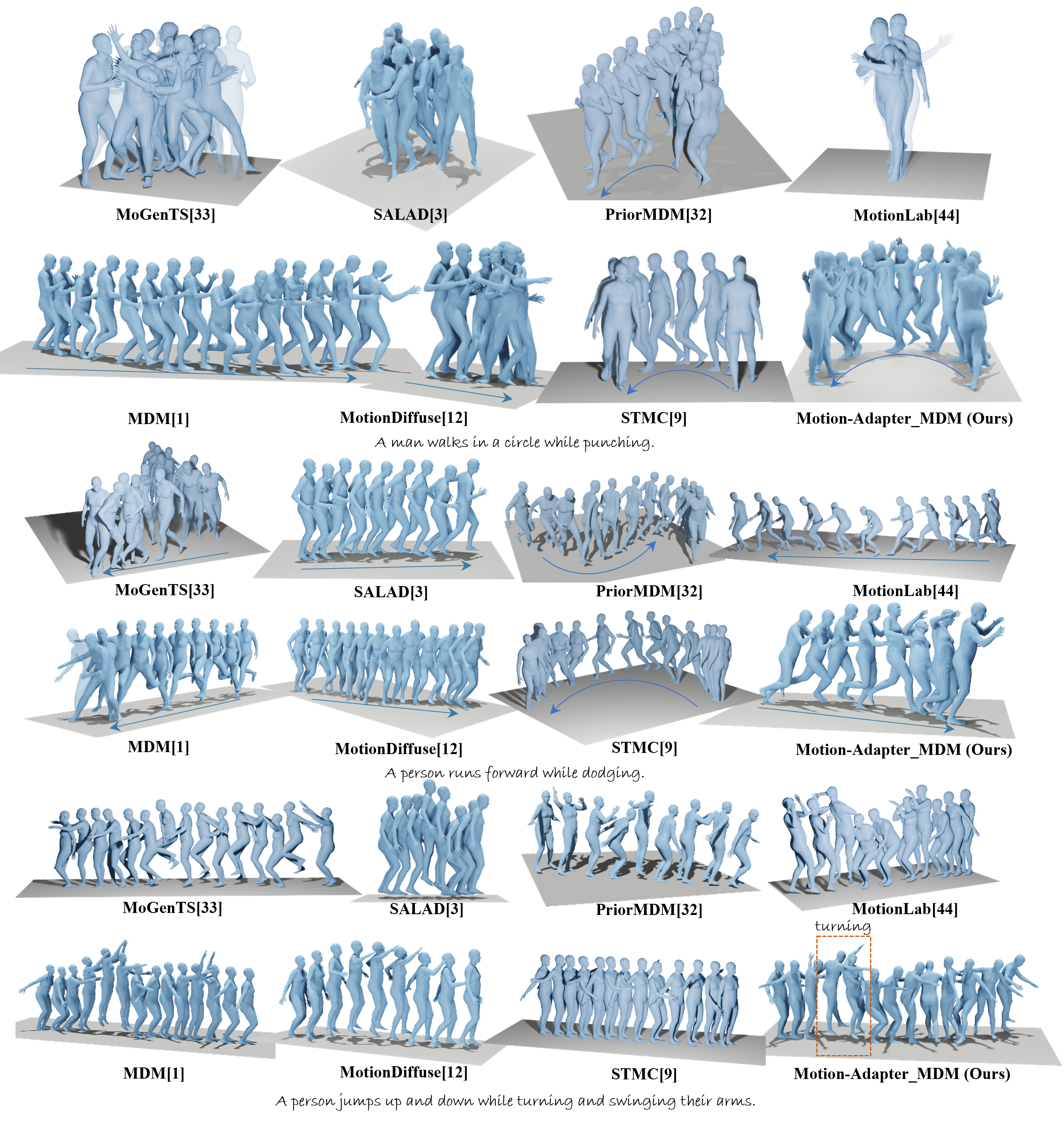}
        \caption{Qualitative comparison results of compound actions with complex textual prompts. To enhance visual clarity, the later frames are rendered with increased transparency, and arrows indicate the direction of motion.}
        \label{fig: LongSentence}
\end{figure*}

In this section, we present examples of complex prompts in Figures \ref{fig: teaser} and ~\ref{fig: LongSentence}, featuring textual descriptions with more detailed action specifications and extended temporal structures. 

\textbf{\textit{Comparison to text-to-motion models}.} As illustrated, similar to results from prompts containing two simple sub-action terms without detailed descriptions, existing methods often produce only a single action or execute two actions sequentially. For example, in Figure \ref{fig: teaser}, for the prompt 'walking backward while lifting dumbbells', methods such as MDM~\cite{HMDM} and SALAD~\cite{SALAD} tend to generate either 'lifting dumbbells' or 'walking backward', respectively. MotionDiffuse~\cite{MotionDiffuse}, while capable of producing the compound action, exhibits unnatural upper-body motion for 'lifting dumbbells'. In example 1 shown in Figure \ref{fig: LongSentence}, most baseline methods generate 'punching' but fail to capture 'walking in a circle', while PriorMDM~\cite{priorMDM} omits 'punching'. Similarly, in the second example, 'dodging' is ignored by methods such as MotionDiffuse~\cite{MotionDiffuse} and SALAD~\cite{SALAD}, whereas MDM~\cite{HMDM} produces 'running' and 'dodging' sequentially. In the third example, although all baseline methods generate 'jumping up and down', the 'turning' component is largely ignored, and MotionLab fails to generate 'swimming.'


\textbf{\textit{Comparison to spatial composition models}.} In these three examples, STMC~\cite{STMC} generates only partial sub-actions, failing to fully represent the compound actions. Specifically, it produces only 'walk in a circle' while omitting 'punching', 'runs forward' without 'dodging', and 'swimming' without the 'jump' and 'turn' components. Overall, STMC~\cite{STMC} demonstrates limited capability in capturing compound actions, as it cannot generate multiple sub-actions simultaneously, resulting in sequences that lack completeness and richness.

\textbf{\textit{Summary.}} From results on complex prompts, it is evident that in scenarios such as 'jump and turn', which require full-body coordination, our approach produces natural motions, demonstrating that the learned masks provide flexible guidance rather than rigid constraints. The turning motion results further indicate that our method robustly captures intricate spatio-temporal dependencies and preserves coherent coordination, even in cases involving overlapping or concurrent movements.

\subsection{Results of Motion-Adapter\_MotionDiffuse}
\label{sec: MotionDiffuseResults}

\begin{figure*}[htb]
        \centering
        \includegraphics[width=\textwidth]{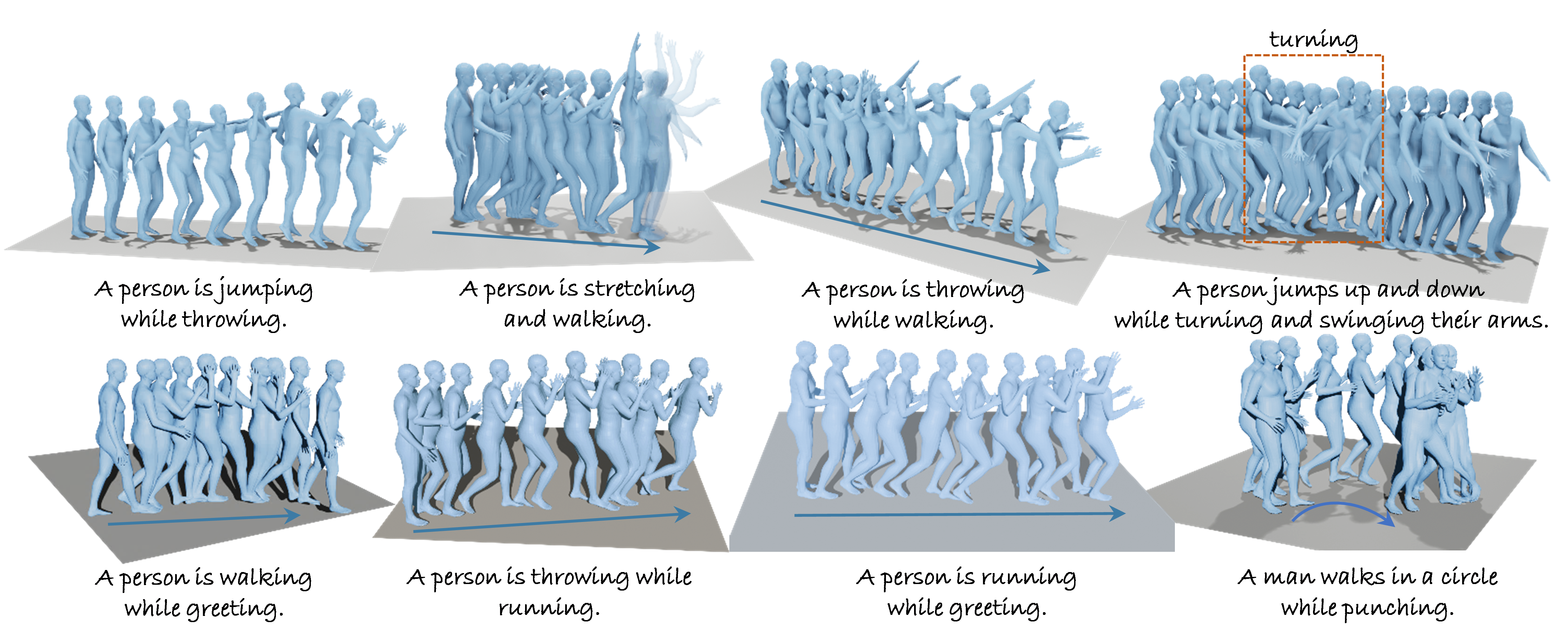}
        \caption{Results of our Motion-Adapter\_MotionDiffuse.}
        \label{fig: Our_MotionDiffuse}
\end{figure*}

The qualitative results of our Motion-Adapter\_MotionDiffuse are presented in Figure \ref{fig: Our_MotionDiffuse} with the same textural prompts tested in the previous sections. It can be seen that our Motion-Adapter\_MotionDiffuse consistently produces natural and coherent motions for both simple and complex prompts in a plug-and-play manner, outperforming both text-to-motion and spatial composition baselines. It notably surpasses MotionDiffuse~\cite{MotionDiffuse} in generating compound actions and also outperforms the text-driven motion editing methods in SALAD~\cite{SALAD}, as illustrated in Figure~\ref{fig: motion_edit}.

The results obtained by both Motion-Adapter\_MDM and Motion-Adapter\_MotionDiffuse demonstrate that the proposed Motion-Adapter effectively enhances motion generation across different backbone networks, improving the model's ability to handle complex, multi-part actions. 

\subsection{Quantitative Comparisons}
\label{sec: quantitatives}

For quantitative evaluation, we follow the metric suite proposed in T2M\cite{TwoStage}, where R-Precision and MM-Dist serve as indicators of semantic alignment between the input text and the generated motions. It is important to note that the evaluation model provided by T2M\cite{TwoStage} was originally trained on simple actions and does not encompass compound motion data. As a result, its evaluation of compound actions and their correspondence with textual descriptions can be biased and, in many cases, unreliable. To overcome this limitation, we randomly select motions from HumanML3D dataset\cite{HumanML} and synthesized a new dataset of 2,652 compound motions, which also incorporates the 484 benchmark motions. This dataset was divided into 80\% for training T2M~\cite{TwoStage}, 10\% for validation to fine-tune the model, and 10\% for evaluation. The retrained evaluation model was then employed to compute R-Precision and MM-Dist, thereby providing a fairer and more reliable measurement of how well the generated motions align with their textual prompts.

In addition, we employ the FID metric to evaluate the quality of generated motions by measuring distributional differences between generated and real motions in the feature space. The Diversity metric is further introduced to quantify the variance within the generated motions, thereby reflecting the range of motion styles captured. Finally, to assess the smoothness of motion sequences, we adopt the Transition metric from STMC\cite{STMC}, which computes the Euclidean distance between poses before and after transition moments.

Following the experimental protocol of T2M\cite{TwoStage}, each experiment is repeated 20 times on our compound motion dataset, and the average results together with their 95\% confidence intervals are reported in Table~\ref{tab: Quatitative}. The results show that both of our models consistently achieve superior performance on compound actions, irrespective of the underlying generative network architecture. In particular, lower MM-Dist scores and higher R-Precision scores indicate a stronger degree of semantic alignment between the generated motions and the input text. Similarly, lower FID scores demonstrate that the feature distribution of the generated motions is more closely aligned with that of real motions. Finally, the Diversity and Transition metrics confirm that the variability and smoothness of our generated motions are most consistent with the ground truth. Taken together, these findings underscore the robustness of our approach in generating compound motions that are both semantically faithful and natural in execution—an area where existing methods often struggle to balance semantic accuracy with motion quality.

Both the quantitative and qualitative results consistently demonstrate the effectiveness of our models. They highlight the ability to capture semantic alignment, fidelity, diversity, and smoothness in compound motions, while the generated sequences further exhibit natural transitions and coherent action composition that closely resemble real human motion. 

\clearpage

\begin{table*}[tb]
  \caption{Quantitative results on the test sets. Symbols: $\uparrow$ (higher is better), $\downarrow$ (lower is better), and $\rightarrow$ (closer to the real distribution is better). Each method was evaluated 20 times; ± indicates the 95\% confidence interval. The best scores are shown in \textbf{bold}, while the second-best scores are \underline{underlined}.}
  
  \label{tab: Quatitative}
  \small
  \begin{tabular}{lcccccc}
    \toprule
    
    \textbf{Methods} & \multicolumn{2}{c}{\textbf{R-Precision}$\uparrow$} & {\textbf{MM-Dist} $\downarrow$} & {\textbf{FID} $\downarrow$} & {\textbf{Diversity} $\rightarrow$} & {\textbf{Transition} $\rightarrow$} \\
 & \textbf{Top-1} & \textbf{Top-3} & & & & \\

    \hline

    Ground Truth & $0.991 ^{\pm.003}$ & $1.000^{\pm.000}$ &  $0.413^{\pm.001}$ & $0.014^{\pm .001}$ & $7.134^{\pm .178}$ &  $1.872^{\pm .017}$ \\

    MDM~\cite{HMDM} & $ 0.121^{\pm.009}$ & $0.282^{\pm.013}$ & $16.132^{\pm .108}$ &  {$8.019^{\pm .189}$} & $8.470^{\pm .066}$ & $2.248^{\pm .028}$   \\

    MotionDiffuse~\cite{MotionDiffuse} &  $0.109^{\pm .006}$ &  $0.253^{\pm .009}$ &  $16.380^{\pm.062}$ & {$4.733^{\pm .056}$} & $8.556^{\pm .086}$ &  $0.250^{\pm .004}$  \\

    PriorMDM~\cite{priorMDM} & $0.076^{\pm .007}$ & $0.174^{\pm .007}$ & $22.350^{\pm .130}$ &  {$15.712^{\pm .759}$} & $8.344^{\pm.244}$ &   $1.194^{\pm.052}$  \\

    MotionLab~\cite{Motionlab} &  $0.141^{\pm .008}$ & $0.338^{\pm .019}$ &  $15.238^{\pm .115}$ & $8.342^{\pm .099}$ &  $8.866^{\pm .068}$ &  $0.017^{\pm .001}$ \\

    MoGenTS~\cite{MoGenTS} & $0.137^{\pm .011}$ &  $0.334^{\pm .016}$  & $14.991^{\pm .068}$ &  $9.510^{\pm .057}$  & $9.306^{\pm .088}$ &  $1.322^{\pm .027}$  \\

        STMC\cite{STMC}  & $0.051^{\pm .007}$ & $0.132^{\pm.009}$ & $19.056^{\pm .043}$ & $52.563^{\pm .067}$ & $1.003^{\pm .036}$ &  $0.012^{\pm .001}$  \\

    SALAD~\cite{SALAD}& $0.154^{\pm .006}$ & $0.357^{\pm  .011}$ & $15.018^{\pm .056}$ &   {$14.363^{\pm .109}$} & $9.572^{\pm .112}$ & $0.366^{\pm .007}$ \\

    SALAD\_Motion Editing~\cite{SALAD}  &  {$0.078^{\pm .007}$} &  {$0.217^{\pm .010}$} &  {$17.723^{\pm .062}$} &  {$27.131^{\pm .247}$} & $5.938^{\pm.070}$ &  $1.096^{\pm.006}$  \\

    \hline
    
    Motion-Adapter\_{MDM} (Ours) & \underline{$0.158^{\pm .008}$} & \boldsymbol{$0.360^{\pm .019}$} & \boldsymbol{$14.952^{\pm .137}$} & \boldsymbol{$ 3.592^{\pm .086}$} & \boldsymbol{$7.744^{\pm .122}$} &  \boldsymbol{$1.936^{\pm .017}$} \\

    Motion-Adapter\_{MotionDiffuse} (Ours) & \boldsymbol{$0.162^{\pm.009}$} & \underline{$0.359^{\pm .012}$} & \underline{$14.988^{\pm .081}$} & \underline{$3.719^{\pm .049}$} & \underline{8.246 $^{\pm .176}$} & \underline{$1.381^{\pm .011}$} \\
    \bottomrule

  \end{tabular}
\end{table*}

\subsection{User Study}
\label{sec: userStudy}

To further evaluate the perceptual quality and fidelity of the generated motions, we conducted a user study involving 65 participants. Each participant was asked to complete three main tasks: (i) rate 15 videos per method based on the semantic alignment between the motion and the corresponding textual prompt, providing a measure of fidelity for the generated results; (ii) watch another 15 videos per method and identify the actions being performed—only the action words were required, to evaluate the perceptual quality of the motions; and (iii) evaluate the overall generation thereby providing an overall measure of motion quality. Tasks (i) and (iii) were rated on a 10-point scale, while for task (ii), we calculated the percentage of correctly identified actions. This design allowed us to comprehensively evaluate both the semantic correctness and the perceptual realism of the generated motions.

The scores, summarized in Table~\ref{tab: UserStudy}, clearly indicate the superiority of Motion-Adapter-based methods. These models achieved the highest scores across all evaluation metrics, with fidelity ratings of 9.27 and 9.08, demonstrating that the majority of generated motions closely align with the corresponding textual prompts. In contrast, all baseline methods scored below 6, highlighting their limitations in capturing compound actions. For task ii, where participants identified actions based solely on observing the motion, the baseline methods achieved an accuracy of approximately 50\%. This reduced performance is largely due to baseline models frequently generating only partial or sequential actions, failing to capture the intended compound actions. Together, these evaluations confirm that Motion-Adapter substantially outperforms existing baselines in terms of semantic fidelity and perceptual clarity.

Interestingly, for overall motion quality (task iii), all methods scored similarly, reflecting their ability to produce visually plausible motions regardless of semantic accuracy. However, this comparison highlights that while baselines may appear reasonable superficially, only Motion-Adapter consistently produces motions that are both semantically accurate and perceptually coherent.

\begin{table} [H]
  \caption{A user study was conducted with 65 participants. PQ denotes Perceptual Quality, and MQ denotes Motion Quality. The best scores are shown in \textbf{bold}, while the second-best scores are \underline{underlined}.}
  
  \label{tab: UserStudy}
  \small
  \begin{tabular}{lccc}
    \toprule
    
    \textbf{Methods} & \textbf{Fidelity} & \textbf{PQ} & \textbf{MQ}\\

    \hline

    MDM\cite{HMDM} & 4.78 & 56.78\% & 8.35 \\
    
    MotionDiffuse\cite{MotionDiffuse}  & 3.75 & 46.87\%  & 7.98 \\

    PriorMDM~\cite{priorMDM} & 5.56 & 61.56\% & 7.84 \\

    MotionLab~\cite{Motionlab} & 4.78 & 54.63\% & 8.13 \\

    MoGenTS~\cite{MoGenTS}  & 4.38 & 42.59\% & 7.82 \\
    
    STMC\cite{STMC} & 4.02 & 57.78\% & 8.17    \\

     SALAD\cite{SALAD} & 4.13 & 41.47\% & 7.78      \\

    SALAD\_Motion Editing \cite{SALAD} & 3.83 & 42.56\% & 7.43      \\
    
    \hline
    Motion-Adapter\_MDM (Ours) & \textbf{9.27} & \textbf{89.67\%} &  \underline{8.91} \\

    Motion-Adapter\_MotionDiffuse (Ours) & \underline{9.08} & \underline{87.34\%} &  \textbf{9.05} \\
    \bottomrule

  \end{tabular}
\end{table}

\subsection{Ablation Study}
\label{sec: Ablation}

In this section, we conduct ablation studies to validate the effectiveness of the proposed method, demonstrating how each design choice contributes to enhancing motion generation. The primary factors under investigation are the Motion-Adapter itself and the masking step constraints.

\textbf{\textit{Removing Motion-Adapter.}} The primary contribution to the generation of compound actions arises from the Motion-Adapter, which extracts attention maps that serve as structural masks. By explicitly guiding the network to disentangle spatial dependencies across different body parts, these masks enable the model to capture the structural relationships underlying complex, multi-part motions. When the Motion-Adapter is removed, the model reduces to the original MDM~\cite{HMDM} or MotionDiffuse~\cite{MotionDiffuse}. Both qualitative and quantitative results shown in Sections \ref{sec: simplePrompts}, \ref{sec: complexePrompts}, \ref{sec: quantitatives}, and \ref{sec: userStudy} demonstrate a clear degradation in motion quality and coherence, underscoring the central role of the Motion-Adapter in generating compound actions.

\begin{table*}[bt]
  \caption{Quantitative results of ablation studies. Symbols: $\uparrow$ (higher is better), $\downarrow$ (lower is better), and $\rightarrow$ (closer to the real distribution is better). Each method was evaluated 20 times; ± indicates the 95\% confidence interval. Best results are in \textbf{bold}.}
  
  \label{tab: ablation}
  \small
  \begin{tabular}{lcccccc}
    \toprule
        \textbf{Methods} & \multicolumn{2}{c}{\textbf{R-Precision}$\uparrow$} & {\textbf{MM-Dist} $\downarrow$} & {\textbf{FID} $\downarrow$} & {\textbf{Diversity} $\rightarrow$} & {\textbf{Transition} $\rightarrow$} \\
 & \textbf{Top-1} & \textbf{Top-3} & & & & \\

    \hline

        Ground Truth & $0.991 ^{\pm.003}$ & $1.000^{\pm.000}$ &  $0.413^{\pm.001}$ & $0.014^{\pm .001}$ & $7.134^{\pm .178}$ &  $1.872^{\pm .017}$ \\

             \hline

         MDM & $ 0.121^{\pm.009}$ & $0.282^{\pm.013}$ & $16.132^{\pm .108}$ &  {$8.019^{\pm .189}$} & $8.470^{\pm .066}$ & $2.248^{\pm .028}$   \\

        Masking step &  $0.142^{\pm .013}$ & $0.311^{\pm .013}$ & $15.444^{\pm .093}$ & $ 3.833^{\pm .078}$ & $7.832^{\pm .274}$ &  $1.972^{\pm .033}$ \\ 
    
    Motion-Adapter\_{MDM} (Ours) & \boldsymbol{$0.158^{\pm .008}$} & \boldsymbol{$0.360^{\pm .019}$} & \boldsymbol{$14.952^{\pm .137}$} & \boldsymbol{$ 3.592^{\pm .086}$} & \boldsymbol{$7.744^{\pm .122}$} &  \boldsymbol{$1.936^{\pm .017}$} \\

     \hline         

         MotionDiffuse &  $0.109^{\pm .006}$ &  $0.253^{\pm .009}$ &  $16.380^{\pm.062}$ & {$4.733^{\pm .056}$} & $8.556^{\pm .086}$ &  $0.250^{\pm .004}$  \\

         Masking step &  $0.152^{\pm .009}$ & $0.339^{\pm .008}$ & $15.468^{\pm .087}$ & $ 4.364^{\pm .174}$ & $8.551^{\pm .338}$ &  $1.416^{\pm .023}$ \\ 

    Motion-Adapter\_{MotionDiffuse} (Ours) & \boldsymbol{$0.162^{\pm.009}$} & \boldsymbol{$0.359^{\pm .012}$} & \boldsymbol{$14.988^{\pm .081}$} & \boldsymbol{$3.719^{\pm .049}$} & \boldsymbol{$8.246^{\pm.176}$} & \boldsymbol{$1.381^{\pm .011}$} \\
    
    \bottomrule

  \end{tabular}
\end{table*}

\textbf{\textit{Masking steps.}} We further validate our decision to extract cross-attention maps before the denoising step $t=750$ and halt employing masks after the denoising step $t=250$.

This decision is based on the observation that the core structure of each action is largely determined during the early denoising steps, which is similar to the case in 2D diffusion models\cite{Attend-Excite}. Consequently, extracting cross-attention maps and employing the masks in the middle or later stages of the process has minimal impact on the final motion. Additionally, in the later denoising steps, the diffusion model focuses on fusing temporal and spatial information to produce coherent motions. At this stage (\textit{i.e.}, $t<250$), we cease applying the mask to preserve the model's ability to generate natural and fluid motion. 

\begin{figure}[H]
\centering
\includegraphics[width=0.5\textwidth]{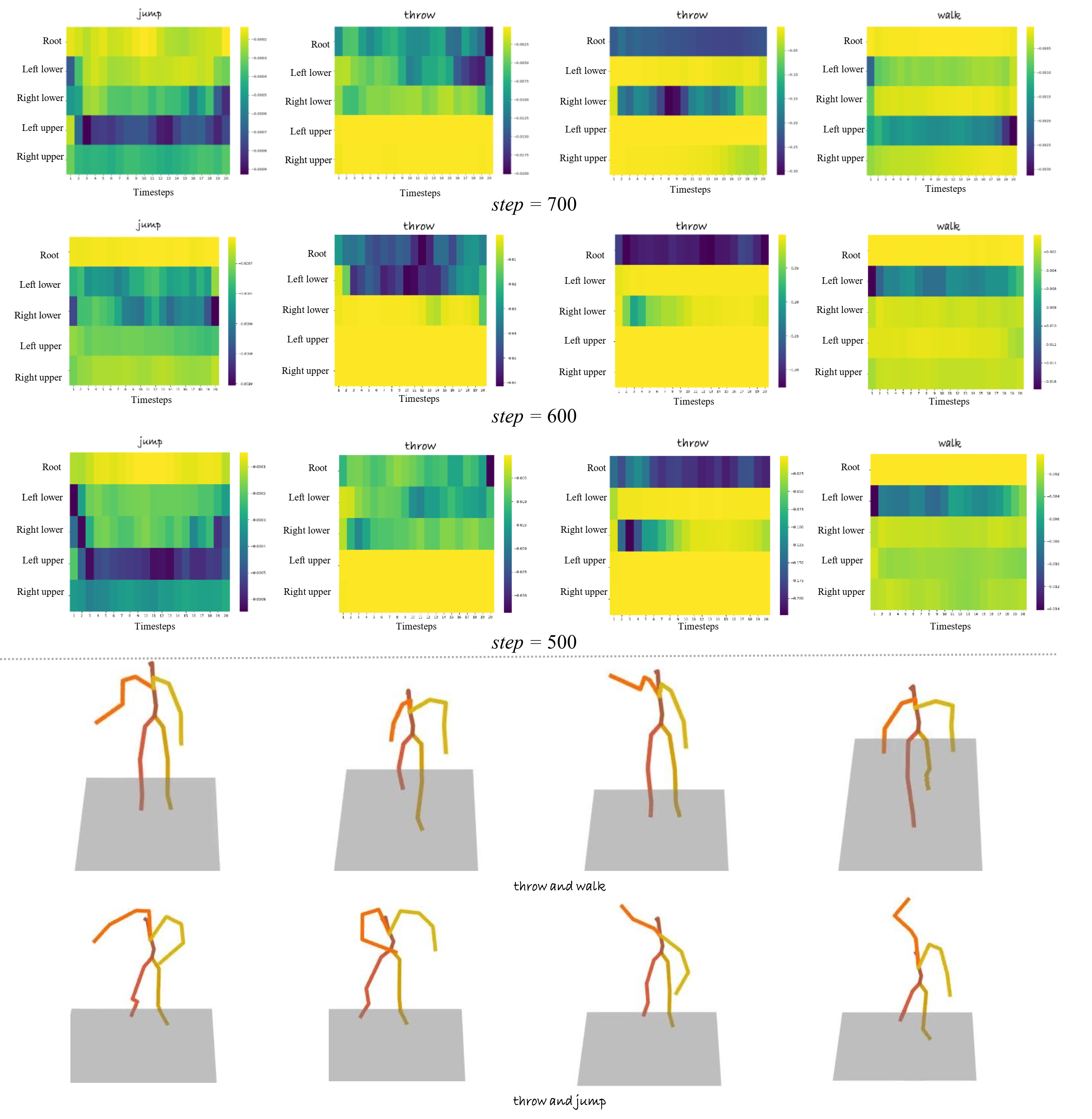}
    \caption{Attention maps extracted at $t = 700, 600, 500$ along with the resulting motion generated by applying the masks throughout all denoising steps. For visual clarity, we show the skeletons.}
    \label{fig: AblationStudySupp}
\end{figure}

As shown in Table~\ref{tab: ablation}, removing the masking step constraints leads to decreased performance across all evaluation metrics, suggesting that unconstrained masking places excessive emphasis on specific joints. We further observe an overall upward trend in the Transition metric, caused by non-smooth transitions that produce abrupt changes at motion boundaries. This phenomenon is also illustrated in Figure~\ref{fig: AblationStudySupp}: the top row shows attention maps extracted when $t < 750$, which appear less informative and frequently misidentify body parts, while the bottom row demonstrates that applying masks at all timesteps results in unnatural motion representations due to insufficient feature fusion. Such over-activation disrupts the coordination between upper and lower body, ultimately degrading the smoothness and realism of the generated motions. Notably, applying the masks generated by the Motion-Adapter improves scores compared to the original MDM\cite{HMDM} and MotionDiffuse\cite{MotionDiffuse}, further validating the effectiveness of the proposed method from an additional perspective.

\section{Limitations}

While the Motion-Adapter effectively enables the synthesis of compound actions, two main limitations remain. First, its generative capacity is bounded by the pre-trained diffusion backbone, as no additional training is performed; our method specifically addresses compound actions but cannot exceed the generative power of the underlying diffusion model. Second, the current design treats the upper and lower body as unified regions, restricting fine-grained control over individual parts such as the hands or fingers—an essential aspect for generating nuanced and expressive motions.

\section{Conclusion}

In this work, we introduce the Motion-Adapter, designed to facilitate the generation of compound action motions. It incorporates a dual-alignment mechanism that enables the generation of decoupled cross-attention maps and the manipulation of noise-level features during the denoising process—all without modifying the backbone architecture, altering its parameters, or requiring additional training data or external conditions.

While our primary focus is on using the Motion-Adapter to address issues such as catastrophic neglect and attention representation collapse, we posit that it can be extended to broader motion editing and generation tasks through the definition of task-specific masks. Moreover, establishing fine-grained correspondences between body parts and textual tokens presents a promising direction for future research, offering more precise and controllable motion generation.

\section{Acknowledgment}

The authors would like to thank the editors and reviewers for their insightful comments. We are also grateful to Daniel Cohen-Or for his valuable suggestions, which helped improve this work. This work was supported by the National Science Foundation of China [62576274], Natural Science Basic Research Program of Shaanxi [2025JC-QYXQ-039], and the Shaanxi Science and Technology Association Youth Talent Support Program [20230115].

\bibliographystyle{IEEEtran}
\footnotesize
\bibliography{ref}

\vspace{-10mm}
\begin{IEEEbiography}[{\includegraphics[width=1in, height=1.25in,clip,keepaspectratio]{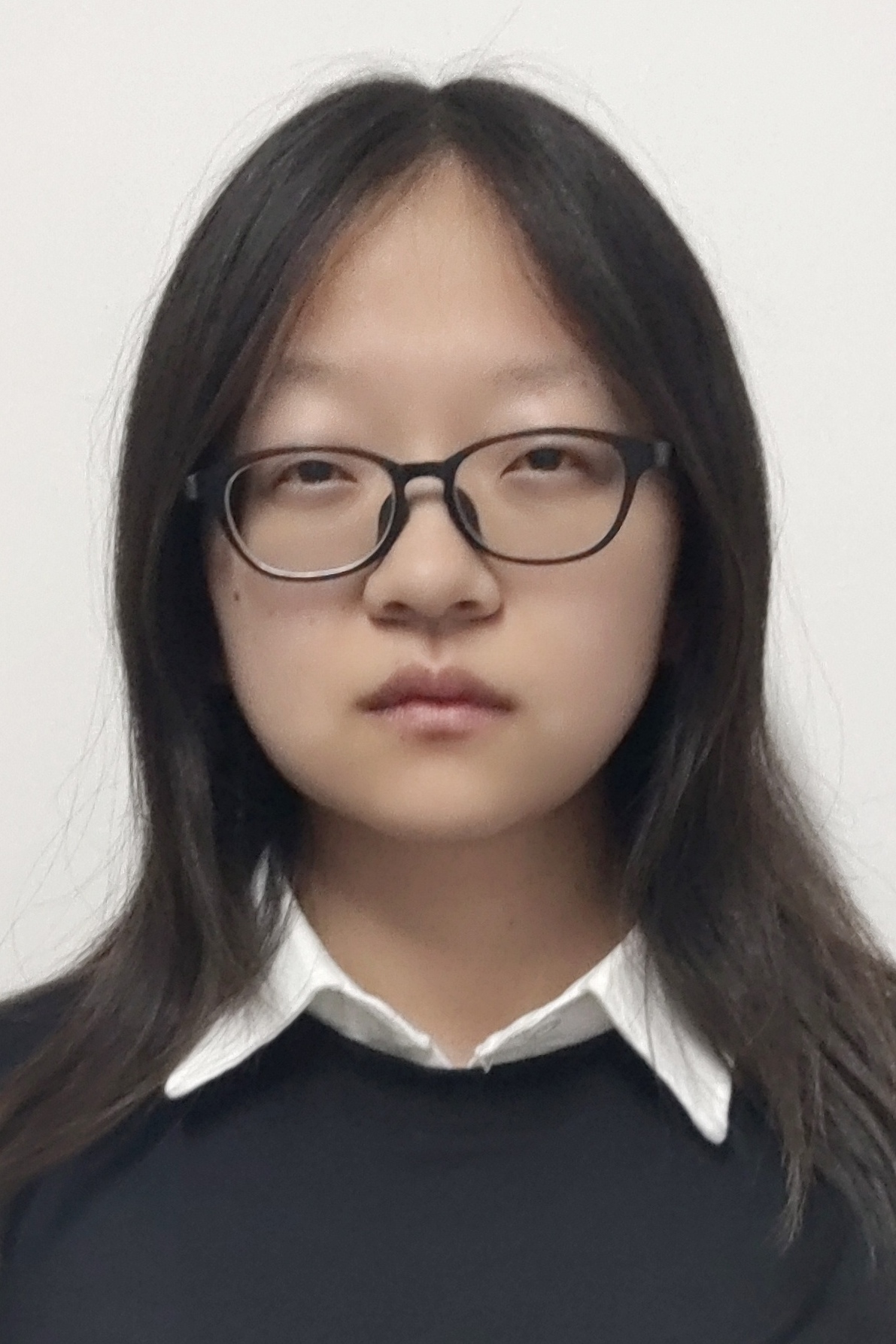}}]{Yue Jiang}
received the B.S. degree in Software Engineering from Northwest University, China, in 2023. Since July 2023, She has been pursuing the M.S. degree in Software Engineering at Northwest University. Her research interests include computer graphics, motion synthesis, and deep learning.
\end{IEEEbiography}
\vspace{-15mm}

\begin{IEEEbiography}[{\includegraphics[width=1in, height=1.25in, clip,keepaspectratio]{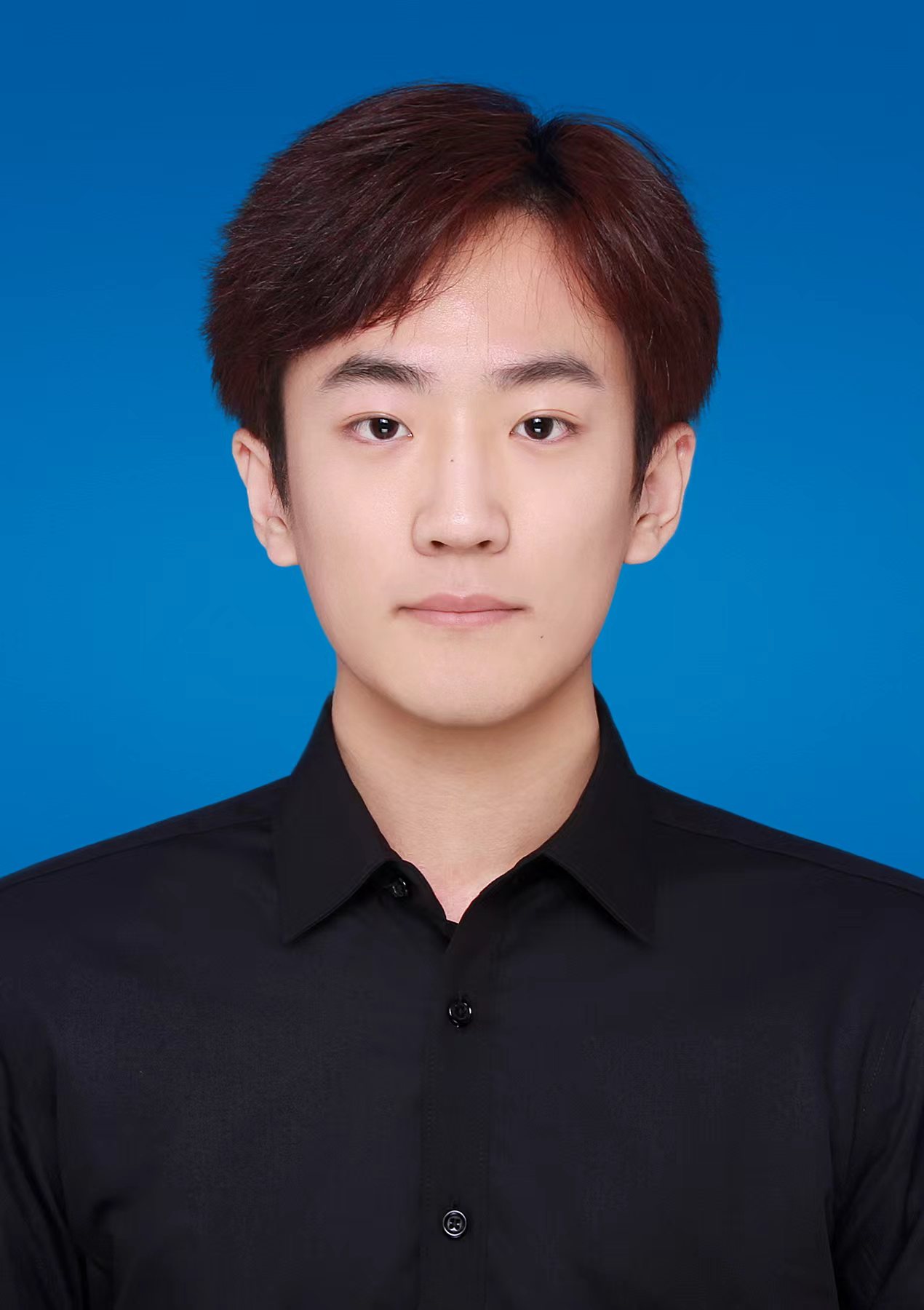}}]{Mingyu Yang} has been pursuing the B.S. degree in Software Engineering at the School of Computer Science, Northwest University of China, since 2022. His research interests include machine learning and artificial intelligence.
\end{IEEEbiography}
\vspace{-15mm}

\begin{IEEEbiography}[{\includegraphics[width=1in, height=1.25in, clip,keepaspectratio]{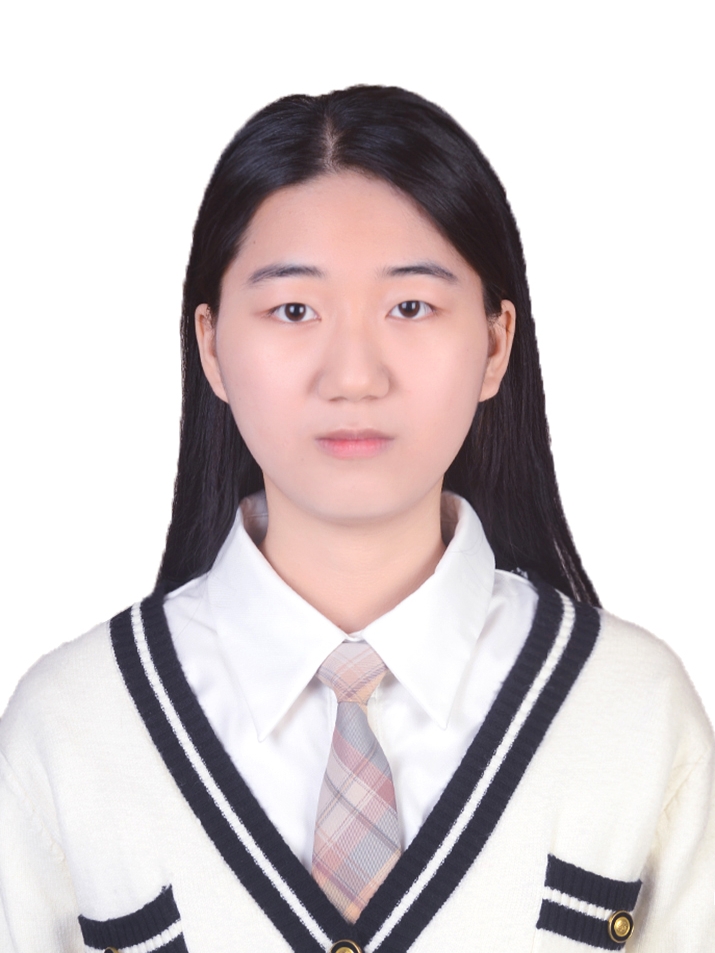}}]{Liuyuxin Yang} received the B.S. degree in Software Engineering from Northwest University, China, in 2023. She is currently working toward the M.S degree in software engineering with the School of Computer Science, Northwest University of China. Her research interests include visualized analysis and deep learning.
\end{IEEEbiography}
\vspace{-15mm}

\begin{IEEEbiography}[{\includegraphics[width=1in, height=1.25in, clip,keepaspectratio]{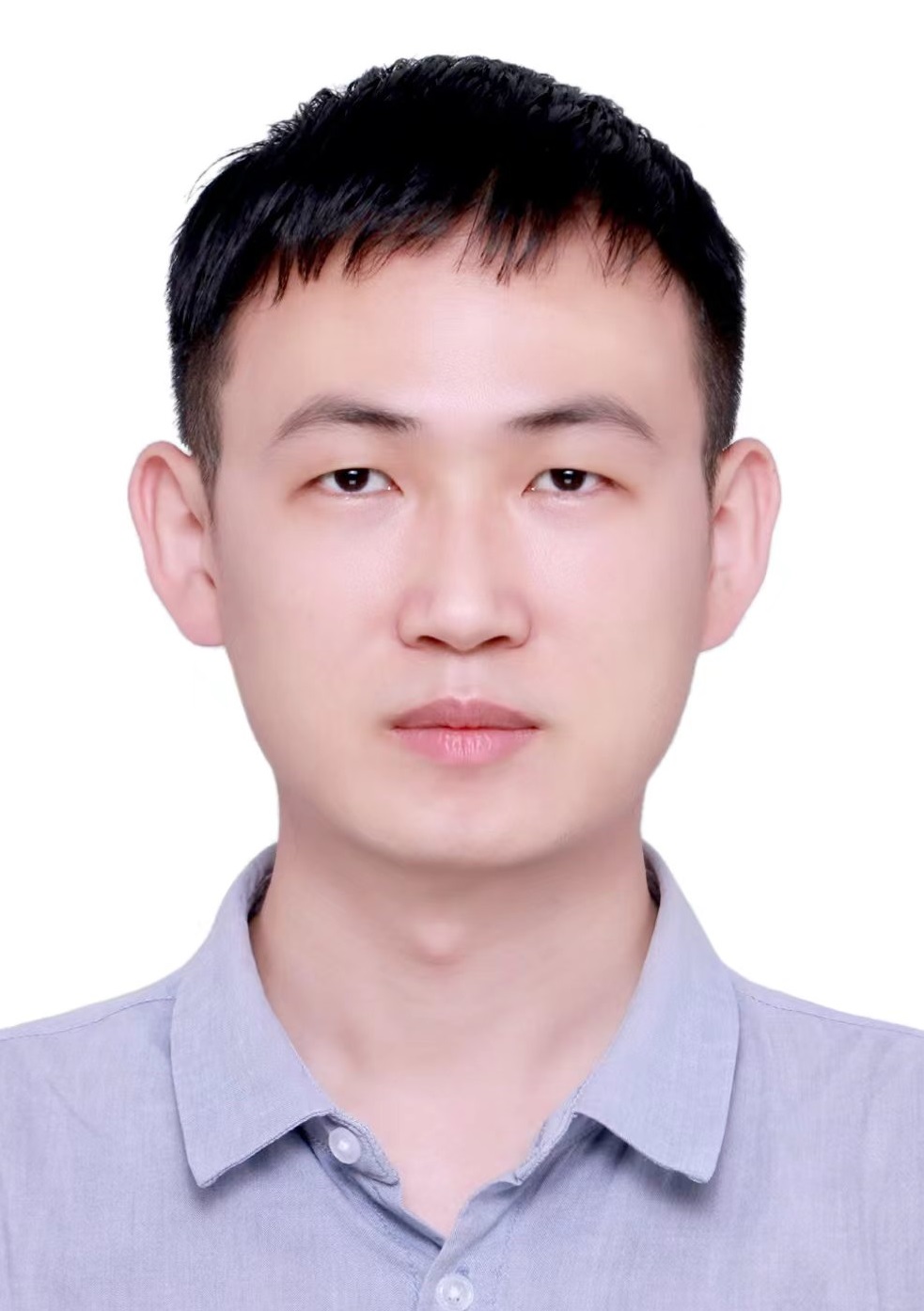}}]{Yang Xu} received his B.E. and Ph.D. degrees from Beihang University in 2014 and 2020, respectively. He is currently an associate professor in the School of Computer Science, Northwest University. His research interests include computer graphics and computer vision.
\end{IEEEbiography}
\vspace{-15mm}

\begin{IEEEbiography}[{\includegraphics[width=1in, height=1.25in, clip,keepaspectratio]{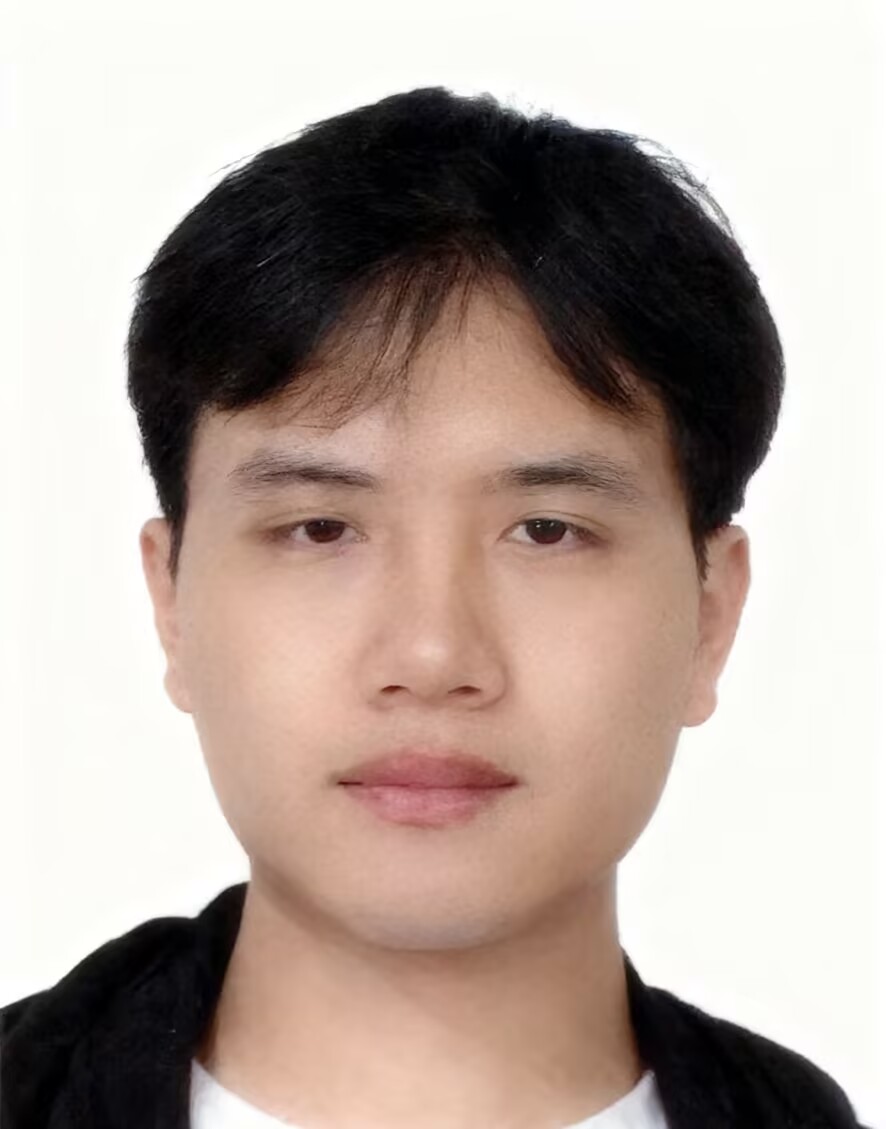}}]{Bingxin  Yun} was admitted to Northwest University in 2023 and is currently pursuing a Bachelor of Science (B.S.) degree in Software Engineering at the university's School of Computer Science. His research interest focuses on deep learning.
\end{IEEEbiography}
\vspace{-15mm}

\begin{IEEEbiography}[{\includegraphics[width=1in, height=1.25in, clip,keepaspectratio]{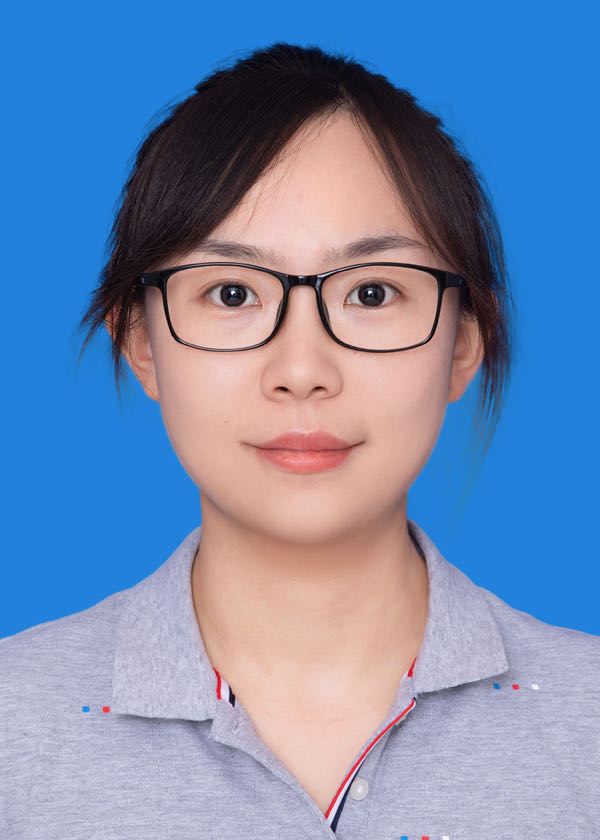}}]{Yuhe Zhang}
(Member IEEE) received the B.S. degree in software engineering and Ph.D. degree in computer applied technology from Northwest University of China in 2012 and 2017. From July 2017 to July 2020, she was a lecturer at School of Information Science and Technology, Northwest University of China. Since August 2020, she has been an Associate Professor at School of Informa- tion Science and Technology, Northwest University of China. Her research interests include computer graphics, image processing, intelligent information
processing and the digital restoration of cultural heritage.
\end{IEEEbiography}

\end{document}